\documentclass{article}

\PassOptionsToPackage{most}{tcolorbox}
\usepackage[final]{acl}

\usepackage{times}
\usepackage{latexsym} 
\usepackage[utf8]{inputenc}
\usepackage[T1]{fontenc}
\usepackage{amsmath,amssymb}
\usepackage{amsfonts}
\usepackage{booktabs}
\usepackage{microtype}
\usepackage{inconsolata}
\usepackage{graphicx}
\usepackage{wrapfig}
\usepackage{enumitem}
\usepackage{pifont}
\usepackage{nicefrac}
\usepackage{listings}

\usepackage{enumitem}
\usepackage{multirow}
\usepackage{wrapfig}
\usepackage{tabularx}
\usepackage{svg}
\usepackage[utf8]{inputenc} 
\usepackage[T1]{fontenc}    
\usepackage{hyperref}       
\usepackage{url}            
\usepackage{booktabs}       
\usepackage{amsfonts}       
\usepackage{nicefrac}       
\usepackage{microtype}      
\usepackage{xcolor}         
\usepackage{placeins}
\usepackage{caption}
\usepackage[most]{tcolorbox}
\usepackage{longtable}
\usepackage{array}
\usepackage{placeins}
\tcbuselibrary{listings,breakable}
\tcbuselibrary{skins,breakable}

\tcbuselibrary{listings, breakable, skins}
\definecolor{DeepPurple}{HTML}{673AB7}
\definecolor{LighterGray}{HTML}{FAFAFA}
\definecolor{CaseOrange}{HTML}{F57C00}
\definecolor{formatblue}{RGB}{51,51,144}
\definecolor{formatlavender}{RGB}{224,224,255}
\definecolor{formatgray}{RGB}{242,242,242}
\definecolor{formatblack}{RGB}{25,25,25}
\definecolor{formatolive}{RGB}{142,140,90}
\definecolor{formatcream}{RGB}{250,248,236}
\definecolor{formatteal}{RGB}{0,94,96}

\newtcblisting{schemaformatbox}[1]{
  enhanced,
  breakable,
  boxrule=0.2pt,
  arc=2pt,
  colback=formatcream,
  colframe=formatolive,
  coltitle=white,
  colbacktitle=formatolive,
  coltext=formatteal,
  title=#1,
  fonttitle=\bfseries,
  listing only,
  listing options={
    basicstyle=\ttfamily\normalsize\color{formatteal},
    breaklines=true,
    columns=fullflexible
  }
}

\newtcblisting{schemabox}[1]{
  enhanced,
  breakable,
  colback=gray!5,
  colframe=black,
  coltitle=white,
  colbacktitle=black,
  title=#1,
  fonttitle=\bfseries,
  listing only,
  listing options={
    basicstyle=\ttfamily\small,
    breaklines=true,
    columns=fullflexible
  }
}

\newcommand{\answerTODO}[1][]{\textcolor{red}{\bf [TODO]}}
\newcommand{\justificationTODO}[1][]{\textcolor{red}{\bf [TODO]}}

\title{HomeFlow: A Data Flywheel for Smart Home Agent Training with Verifiable Simulation
}

\author{%
  \textbf{Yi Gu$^{1,*}$}\quad
  \textbf{Huacan Wang$^{1,*,\dagger,\ddagger}$}\quad
  \textbf{Shuo Zhang$^{1,2,*,\diamond}$}\quad
  \textbf{Yuqing Hou$^{1}$}\\
  \textbf{Lei Xue$^{1}$}\quad
  \textbf{Weipeng Ming$^{1}$}\quad
  \textbf{Chen Liu$^{1}$}\quad
  \textbf{Fangzhou Yu$^{1,3,\diamond}$}\\
  \textbf{Kuan Li$^{1}$}\quad
  \textbf{Ronghao Chen$^{4}$}\quad
  \textbf{Sen Hu$^{4}$}\quad
  \textbf{Xiaofeng Mou$^{1}$}\quad
  \textbf{Yi Xu$^{1,\dagger}$}\\[6pt]
$^{1}$Midea Group\quad
$^{2}$Beijing University of Posts and Telecommunications\\
$^{3}$Donghua University\quad
$^{4}$Peking University\\[3mm]
$^{*}$Equal Contribution,\quad
$^{\dagger}$Corresponding Author,\quad
$^{\ddagger}$Project Leader.\\[2mm]
$^{\diamond}$Work completed during an internship at Midea AI Research Center.
}

\begin{document}

\maketitle


\begin{abstract}
Large language model agents are moving beyond text-only interaction toward physical-world control, with smart homes as a representative domain. Real domestic interaction requires understanding ambiguous intents, operating in dynamic environments, and performing multi-turn reasoning. However, existing methods struggle to generate high-quality training data for smart home agents. We propose \textbf{HomeFlow}, a verifiable data flywheel for this domain. HomeFlow uses HomeEnv as a unified simulation environment and HomeMaker to procedurally generate diverse home settings. Subsequently, Blueprint compiles open-ended user intents into executable state-based success conditions, while MCTS-Flow synthesizes diverse, verifiable multi-turn trajectories through environment-guided tree search. We then optimize the agents via supervised fine-tuning and step-wise RLVE, which facilitates iterative improvement through authentic physical feedback. We further construct SmartHome-Bench to evaluate the agent across various smart home tasks. On this benchmark, HomeFlow-RL-4B and HomeFlow-RL-8B achieve task success rates of 84.60\% and 87.03\%. It is worth noting that HomeFlow-RL-8B even surpasses the leading GPT-5.5 by 1.23 percentage points.

\vspace{0.8em}

\vspace{0.4em}
\noindent\textbf{Correspondence:} 
\href{mailto:wanghuacan17@mails.ucas.ac.cn}{Huacan Wang},
\href{mailto:xuyi42@midea.com}{Yi Xu}\\
\vspace{0.2em}
\noindent\textbf{Code:} Coming soon.
\end{abstract}

\vspace{5mm}

\section{Introduction}

Large language model (LLM)-based agents~\cite{li2025mimovlmiloco} are increasingly moving beyond text-only interaction~\cite{singh2025gpt5, comanici2025gemini25} toward control in the physical world~\cite{xi2025llmagents}. Smart homes~\cite{hammi2022smarthome, huda2024smarthomecities} provide a natural testbed for this transition: they are rich in connected devices, governed by physical states, and closely aligned with everyday user needs. A capable smart home agent must infer implicit user intents, act on the basis of home states, and seek clarification when needed. For example, ``\textit{it feels stuffy}'' may require adjusting the air conditioner or ventilation depending on the current environment. However, directly deploying frontier LLMs as smart home agents is impractical due to their lack of situated awareness. This often leads to seemingly plausible but unexecutable commands, requiring domain-specific training to align models with smart home dynamics.

Such training depends on large-scale multi-turn interaction data, which is difficult to collect from real households due to cost, deployment complexity, and privacy constraints. A common alternative is to synthesize dialogues offline through LLM-based role play~\cite{shao2023characterllm, wang2025coser}. However, text-only synthesis has three fundamental limitations. First, it lacks an underlying physical environment against which actions can be verified. For instance, whether a generated command like ``\textit{lowering the air conditioner}'' maps to a valid device and is actually executable in the current room cannot be determined from text alone, forcing evaluation to rely on subjective LLM-as-judge~\cite{li2024llmjudge, gu2024llmjudge} signals. This introduces noise and prevents reinforcement learning from receiving physically grounded rewards. Second, linear role play typically produces a single trajectory for each scenario, offering limited coverage of diverse environmental evolutions and user responses. Third, training on static offline dialogues mainly supports single-step alignment under fixed contexts~\cite{guo2025deepseekr1, yu2025dapo, zhang2025rlreasoningsurvey}. Without explicit multi-turn training, agents struggle to track intent and correct their policies over full episodes.

To address these limitations, we propose \textbf{HomeFlow}, a verifiable data flywheel for training smart home agents. Its core idea is to construct an interactive simulation environment, HomeEnv, serving as a unified grounded signal source that bridges data generation, trajectory expansion, and multi-turn policy optimization. Built upon HomeEnv, HomeMaker is designed to procedurally generate diverse home layouts, devices, and environmental states, populating the simulation environment. For verifiable data generation, we introduce a pipeline comprising Blueprint and MCTS-Flow. Specifically, Blueprint constructs comprehensive interaction scenarios that incorporate diverse user profiles and contexts, and compiles implicit intents into executable logical constraints. Guided by these blueprints, MCTS-Flow advances data synthesis from one-shot linear sampling to tree-structured search, exploring and verifying diverse multi-turn trajectories. 

Finally, we adopt a two-stage training paradigm. After an initial supervised fine-tuning phase, the agent is further optimized through Step-wise Reinforcement Learning from Verifiable Execution (RLVE)~\cite{zhang2025rlreasoningsurvey}. Unlike traditional approaches that only reward the final outcome, this strategy continuously optimizes the agent during online interactions with a dynamic LLM user and HomeEnv, applying physically grounded rewards at every step of the generation process. 

To evaluate the agent's ability to interact with users and operate devices in realistic smart home scenarios, we construct SmartHome-Bench on top of HomeEnv. On this benchmark, HomeFlow-RL-4B and HomeFlow-RL-8B, trained from Qwen3-4B and Qwen3-8B~\cite{yang2025qwen3} using HomeFlow, achieve overall task success rates of 84.60\% and 87.03\%. Notably, HomeFlow-RL-8B outperforms GPT-5.5 by 1.23\%, while HomeFlow-SFT-4B and HomeFlow-SFT-8B improve over the corresponding text-only role-play fine-tuning baselines by 5.88\% and 4.67\%, respectively. 

\section{Related Work}

\textbf{Smart Home Agents.} 
LLM have emerged as central orchestrators for smart homes, grounding natural language intents into executable API calls~\cite{rivkin2023sage,yu2026iotgpt,birkmose2025homeassistantllm}. Beyond early systems~\cite{king2024sasha,yin2025harmony} focusing on basic instruction decomposition, recent works~\cite{yu2026iotgpt,huang2025homellama,zhan2026hearthnet} explore personalization, persistent memory, and edge-based multi-agent orchestration. Concurrently, benchmarks such as HomeBench~\cite{li2025homebench} and SimuHome~\cite{seo2026simuhome} have significantly driven the development of this field by assessing LLMs on instruction-to-call generation, temporal reasoning, and environment-aware execution.

\textbf{Dialogue Data Synthesis.} Synthetic data generation provides a scalable approach for model post-training. Methods like Self-Instruct~\cite{wang2022selfinstruct} and Evol-Instruct~\cite{xu2023wizardlm} bootstrap and iteratively rewrite instructions, while AgentInstruct~\cite{mitra2024agentinstruct} extends this to complex agentic workflows. For dialogue agents, multi-turn trajectories are typically constructed using user simulators and role-playing, often filtered by LLM-as-a-Judge~\cite{li2024llmjudge,gu2024llmjudge,ni2026usersimulation}. Recent search-based methods~\cite{gao2025websynthesis,koh2024treesearchagents} further integrate tree-structured exploration to enhance trajectory diversity and quality for downstream training.


\textbf{Reinforcement Learning from Verifiable Environments.} While RLHF~\cite{ouyang2022instructgpt} and RLAIF~\cite{bai2022constitutionalai,lee2024rlaif} rely on preference-based rewards, recent trends favor verifiable rewards (RLVR) where correctness is automatically checkable, showing strong potential in mathematics (e.g., GRPO~\cite{shao2024deepseekmath}) and interactive agent environments~\cite{jimenez2023swebench,wei2025webagentr1}. Building on these foundations, HomeFlow formulates smart home task as a stateful learning problem. It leverages verifiable data synthesis to provide a robust cold start, followed by step-wise RLVE optimized through structured, environment-based rewards.
\section{Method}

\begin{figure}[t]
  \centering
  \includegraphics[width=0.9\textwidth]{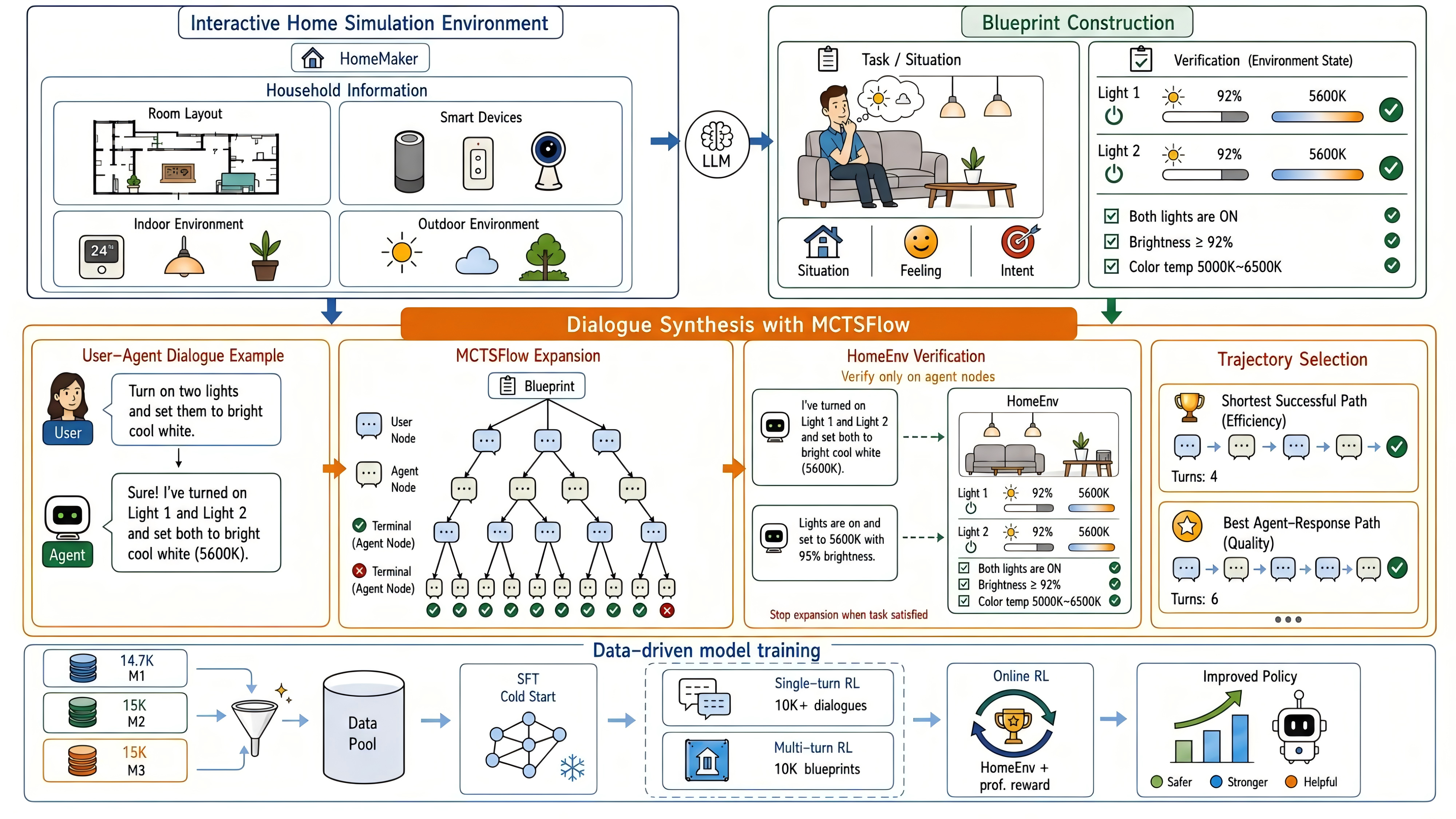}
  \caption{Overview of HomeFlow. Based on HomeEnv, HomeMaker generates diverse home environments. Blueprint then creates executable scene conditions, followed by MCTS-Flow to synthesize diverse, verifiable trajectories. From this architecture, we derive three data synthesis strategies: (M1) pure-text role-play (w/o Blueprint); (M2) Blueprint-grounded linear generation (w/o MCTS-Flow); and (M3) the complete HomeFlow pipeline.}
  \label{fig:main}
\end{figure}

\subsection{Task Definition}
Smart A tasks can be viewed as goal-directed sequential decision making in a partially observable physical environment. We formalize the task as a goal-conditioned Partially Observable Markov Decision Process (POMDP) in discrete time, represented by the tuple $\langle \mathcal{S}, \mathcal{A}, \mathcal{E}, \mathcal{O}, \mathcal{R} \rangle$.

The state space $\mathcal{S}$ captures the underlying physical state of the home. At time step $t$, the state $s_t \in \mathcal{S}$ encodes continuous environmental and device attributes (e.g., temperature), alongside discrete device states (e.g., on/off, mode). To enable bidirectional interaction with both the environment and the user, the agent acts in a heterogeneous action space$\mathcal{A} = \mathcal{A}_{\mathrm{NL}} \cup \mathcal{A}_{\mathrm{Exec}}$, where $\mathcal{A}_{\mathrm{NL}}$ denotes natural language actions used for responses and clarification, and $\mathcal{A}_{\mathrm{Exec}}$ denotes executable control actions, such as generating Python script snippets or API calls to query or modify the physical state.

The environment dynamics are governed by the transition function $\mathcal{E}: \mathcal{S} \times \mathcal{A} \rightarrow \mathcal{S}$. An executable action $a_t \in \mathcal{A}_{\mathrm{Exec}}$ induces a deterministic physical state transition according to $s_{t+1} = \mathcal{E}(s_t, a_t)$, whereas a natural language action $a_t \in \mathcal{A}_{\mathrm{NL}}$ only advances the dialogue context, leaving the physical state unchanged (i.e., $s_{t+1} = s_t$). Because the agent lacks direct access to the full state $s_t$, it makes decisions based on partial observations $o_t \in \mathcal{O}$. The observation $o_t$ may include the user utterance, device-query results, and feedback from executed actions.



To avoid relying on subjective LLM-based evaluation, we represent the user's task goal as a set of verifiable conditions $\Phi = \{\phi_1, \phi_2, \ldots, \phi_K\}$, where each $\phi_k: \mathcal{S} \rightarrow \{0,1\}$ is a boolean predicate over the environmental state. Based on this, given an initial state $s_0 \sim p(s_0)$ and a goal specification $\Phi \sim p(\Phi)$, the agent policy $\pi_\theta(a_t \mid h_t)$ acts upon the interaction history $h_t = (o_1, a_1, \ldots, o_t)$ to generate a complete trajectory $\tau$. Upon reaching the final state $s_T$ at the end of the episode, the terminal verification reward $\mathcal{R}$ is defined by the cumulative satisfaction of all target conditions as $\mathcal{R}(s_T, \Phi) = \sum_{k=1}^{K} \phi_k(s_T)$. This additive formulation provides a denser reward signal by granting partial credit for incomplete tasks. Therefore, the overall objective is to find the optimal policy $\pi_\theta^*$ that maximizes the expected verification success rate:
\begin{equation}
J(\theta) = \mathbb{E}_{\substack{s_0 \sim p(s_0), \, \Phi \sim p(\Phi) \\ \tau \sim p(\tau \mid s_0, \Phi, \pi_\theta)}} \left[ \mathcal{R}(s_T, \Phi) \right]. 
\end{equation}

This formulation highlights two core challenges in training smart home agents: (1) building an interactive execution engine to simulate the environment and compute verifiable rewards (Section~\ref{sec:3.2}); and (2) compiling open-ended scenarios into precise goals to drive efficient exploration and subsequent policy optimization (Section~\ref{sec:3.3} and Section~\ref{sec:3.4}).

\subsection{Interactive Home Environment}
\label{sec:3.2}

We develop HomeEnv, an interactive environment designed for scalable agent training. To diversify the initial state distribution $p(s_0)$, we further introduce HomeMaker, a procedural environment generator. HomeEnv and HomeMaker provide the core environment foundation for HomeFlow.

\subsubsection{HomeEnv: Smart Home Simulation Engine}
HomeEnv serves as the verifiable execution platform throughout data generation and policy optimization. It provides unified execution feedback, partial observations, and rule-based verification. Its design comprises three main components. More details are provided in Appendix~\ref{app:homeenv}.

\textbf{Structured state representation.} We instantiate the home state $s_t \in \mathcal{S}$ as
\begin{equation}
s_t = \langle \mathcal{L}, \mathcal{D}, \psi, \mathcal{X}_t, \mathcal{F} \rangle,
\end{equation}
where $\mathcal{L}$ denotes the set of rooms and their spatial organization, $\mathcal{D}$ is the set of smart devices, and $\psi: \mathcal{D} \rightarrow \mathcal{L}$ maps each device to its location. For each device $d \in \mathcal{D}$, $\mathcal{X}_t(d)$ captures its dynamic states, while $\mathcal{F}(d)$ specifies its exposed executable functions. This declarative device schema decouples state representation from simulation implementation, enabling the environment to scale across heterogeneous and long-tail IoT devices.

\textbf{Execution sandbox.} HomeEnv provides an isolated sandbox for executing actions $a_t \in \mathcal{A}_{\mathrm{Exec}}$. When the agent generates code or API calls, the sandbox returns observations $o_t$ restricted to the direct execution results—such as serialized device states, execution feedback, and runtime errors. This design enforces partial observability, requiring the agent to actively query the environment and interpret feedback. Crucially, it yields high-fidelity interaction traces grounded in actual execution, providing a reliable data source for policy optimization.

\textbf{Verification and auditing module.} This module continuously tracks the state and evaluates the goal conditions $\Phi$ after each step. When all conditions are satisfied, the environment terminates the episode and returns a verified success signal. Additionally, a runtime action auditor blocks unsafe or task-irrelevant operations (e.g., accessing non-existent functions or setting attributes beyond physical limits), converting them into deterministic negative feedback. This dual mechanism establishes a reliable feedback boundary for both offline data filtering and online policy optimization.

\subsubsection{HomeMaker: Diverse Environment Instantiation}
\label{sec:3.2.2}

Smart home environments vary significantly across households in terms of layouts, device configurations, and environmental conditions. Training on a single, fixed environment inevitably leads to severe overfitting. To mitigate this, HomeMaker procedurally generates diverse and physically consistent initial states $s_0$.

For each home instance $i$, HomeMaker first samples real-world weather data to establish the external context. It then constructs a spatial layout $\mathcal{L}_i$ featuring nested structures (e.g., bedrooms with ensuite bathrooms). Based on room semantics, the engine samples devices from prior distributions to determine the device set $\mathcal{D}_i$ and their locations $\psi_i$. Finally, it integrates the external weather, layout structure, and device configurations to initialize the dynamic state variables $\mathcal{X}_0^{(i)}$. This hierarchical process ensures each episode begins with a realistic and logically sound state $s_0^{(i)}$. By expanding the context space through diverse layouts and device combinations, HomeMaker effectively enhances agent robustness across long-tail residential scenarios.

\subsection{Verifiable Data Synthesis}
\label{sec:3.3}

As illustrated in Figure~\ref{fig:main}, leveraging HomeEnv for verifiable execution, we generate large-scale, high-quality training data through a two-stage pipeline: a static Blueprint Construction phase to construct the goal distribution $p(\Phi)$, and a dynamic MCTS-Flow search phase to efficiently explore interaction trajectories $\tau$ under the guidance of environmental verification signals.

\subsubsection{Blueprint Construction}

To synthesize diverse training data at scale, we propose the Blueprint Construction mechanism to orchestrate comprehensive interaction scenarios (i.e., "scripts"). Formally, we define a complete interaction blueprint as a tuple $\mathcal{B} = \langle \mathcal{U}, s_0^{(i)}, \mathcal{I}, \Phi_\mathcal{B} \rangle$. Here, $\mathcal{U}$ represents a multi-dimensional user profile (encompassing physiological characteristics, lifestyle habits, and environmental preferences, e.g., "an elderly person with poor vision in a duplex villa late at night"); $s_0^{(i)}$ is the initial physical state instantiated by HomeMaker in Section~\ref{sec:3.2.2}; $\mathcal{I}$ denotes the user slice-of-life segments generated based on the physical living scenario; and $\Phi_\mathcal{B}$ is the corresponding set of objective evaluation criteria.

Specifically, GPT-5~\cite{singh2025gpt5} acts as a scenario-scripter that, given the initialized physical environment $s_0^{(i)}$ and the user profile $\mathcal{U}$, generates continuous ``slices of life'' of $\mathcal{I}$ within this living space$s_0^{(i)}$. Subsequently, the continuous "slices of life" within $\mathcal{I}$ are segmented into multi-step atomic intents, each formulated as a candidate Boolean evaluation function $\phi_k$. To guarantee absolute objectivity and validity, these generated conditions undergo a rigorous triple-filtering mechanism. First, an \textit{Executability Check} is performed via the HomeEnv sandbox to intercept illegal conditions containing parsing errors or undefined device entities. Second, \textit{Boundary Validity} is verified against the underlying Device Schema to ensure parameters remain within permissible hardware ranges. Finally, the logical consistency between the formulated conditions, the instruction intent, and the user profile is cross-verified using GPT-5. Conditions failing any check are directly discarded. Through this mechanism, ambiguous instructions are strictly reduced into a precise, machine-verifiable goal distribution $p(\Phi)$.

\subsubsection{MCTS-Flow: Verifiable Trajectory Exploration}

With the blueprint $\mathcal{B}$ established, the objective is to synthesize high-quality interaction trajectories $\tau$ that navigate the agent from $s_0^{(i)}$ to a final state satisfying the defined criteria $\Phi_\mathcal{B}$. Traditional linear dialogue generation approaches suffer from severe computational waste: a failure in the final turn of a multi-turn interaction necessitates discarding all preceding exploration. To overcome this problem, MCTS-Flow treats trajectory synthesis as a tree search problem, deeply integrating Monte Carlo Tree Search (MCTS) with HomeEnv environmental verification to efficiently generate diverse, high-quality trajectories.

MCTS-Flow constructs a heterogeneous dialogue search tree $\mathcal{T}=(\mathcal{V},\mathcal{E})$ within the constraints of the blueprint $\mathcal{B}$. Each node $v \in \mathcal{V}$ represents a reusable interaction prefix $s_v = \langle h_v, r_v \rangle$, where $h_v$ is the multi-turn historical context comprising both conversational history and environmental feedback, and $r_v \in \{\text{User}, \text{Agent}\}$ indicates the active role. Three core modifications are introduced to tailor standard MCTS for the specific complexities of smart home scenarios:

\textbf{Role-Aware Adaptive Branching.} We observe that user utterances are strictly constrained by the blueprint intent, whereas agent decisions involve complex intent clarification and multi-step reasoning. Consequently, MCTS-Flow introduces an \textit{asymmetric branching factor} $k(v)$ to manage these distinct roles. Specifically, we allocate a larger sampling width to agent nodes such that $k_{\text{agent}} > k_{\text{user}}$. This strategy concentrates the search budget on the agent's critical decision bifurcations, avoiding redundant sampling of the deterministic user inputs.

\textbf{UCB Selection and Prefix Reuse.} During the selection phase, the Upper Confidence Bound (UCB) algorithm is utilized to balance exploration and exploitation among available child nodes:
\begin{equation}
v^* = \arg\max_{u \in Ch(v),\, \rho(u)=1} \left[ \frac{Q(u)}{N(u)} + c_{\text{ucb}}\sqrt{\frac{2\ln N(v)}{N(u)}} \right] 
\end{equation}
where $Ch(v)$ is the set of child nodes, $N(u)$ is the visit count, $Q(u)$ is the accumulated verification reward, and $\rho(u) \in \{0, 1\}$ is a passable indicator. When a rollout ends, MCTS-Flow will keep all the simulated nodes and mark $\rho(u) = 0$ during backpropagation for all the nodes on the path if all of its child nodes are not passable. This mechanism transforms verified prefixes into reusable assets.

\textbf{Environment-Verified Reward Backpropagation.} Unlike most tree searches relying on model-based scoring, MCTS-Flow leverages the ground-truth feedback provided by the HomeEnv environment. When a rollout reaches a terminal node $v_L$, the fulfillment of the final state $s_T$ is calculated:
\begin{equation}
 R(v_L) = \mathbf{1}\left[ \text{valid}(h_{v_L}) \land \prod_{k=1}^K \phi_k(s_T) = 1 \right] 
\end{equation}
where $\text{valid}(h_{v_L})$ is a binary indicator ensuring the interaction is free of execution errors or command violations. This authentic physical reward is then backpropagated to update the $Q(v)$ and $N(v)$ of all ancestor nodes. Notably, this mechanism not only selects verified trajectories for SFT but also quantifies blueprint complexity through the verified success rate, effectively facilitating curriculum-driven reinforcement learning for increasingly challenging tasks.

\subsection{Two-Stage Agent Training}
\label{sec:3.4}


\subsubsection{Supervised Fine-Tuning}

We extract successful paths from the MCTS-Flow search tree. For each terminal node with verified return $R(v_L)=1$, the corresponding path is converted into a sequential interaction trajectory
$\tau^* = (o_1, a_1, o_2, a_2, \ldots, o_T, a_T)$.
All successful trajectories form the supervised dataset $\mathcal{D}_{\mathrm{SFT}}$. The policy $\pi_\theta$ is optimized with the standard causal language modeling objective:
\begin{equation}
\mathcal{L}_{\mathrm{SFT}}(\theta)
=
-
\mathbb{E}_{\tau^* \sim \mathcal{D}_{\mathrm{SFT}}}
\left[
\sum_{t=1}^{|\tau^*|}
\log \pi_\theta(a_t \mid h_t, \Phi_{\mathcal{B}})
\right].
\end{equation}
SFT injects critical behavioral priors into the agent, including intent clarification, device control semantics, and reasoning over environment observations. However, mere imitation leaves it vulnerable to compounding errors during long-horizon domestic interactions.

\subsubsection{Step-wise RLVE: Aligning Policy with Physical Reality}

In realistic smart homes, user intents and household states co-evolve, rendering offline RL on static conversational prefixes inadequate. To overcome the limitations of passive imitation, we introduce Step-wise RLVE, which optimizes the agent directly through online, multi-turn interactions.

\textbf{Dynamic Trajectory Generation.} During each training episode $e$ initialized from a blueprint $\mathcal{B}$, a strong LLM serves as a dynamic user simulator $\pi_{\text{user}}$. The agent policy $\pi_\theta$ engages in a live interaction loop with both the simulator and the HomeEnv environment, generating a complete online trajectory $\tau^{(e)} = \big(u_1^{(e)}, a_1^{(e)}, r_1^{(e)}, \ldots, u_T^{(e)}, a_T^{(e)}, r_T^{(e)}\big)$, where the user utterance $u_t^{(e)} \sim \pi_{\text{user}}(\cdot \mid \mathcal{B}, h_{<t}^{(e)})$ is dynamically synthesized based on the blueprint constraints and the real-time interaction history. Correspondingly, the agent action $a_t^{(e)} \sim \pi_\theta(\cdot \mid h_{<t}^{(e)}, \Phi_\mathcal{B})$ triggers environmental updates, yielding the step-wise verification reward $r_t^{(e)}$ from Home. Crucially, this formulation ensures that the same blueprint $\mathcal{B}$ yields highly diverse user query sequences across different episodes. Consequently, the agent is forced to optimize over non-stationary input distributions, learning to adapt as the conversational context and household states co-evolve.

\textbf{Composite Verifiable Reward.} Because the environment updates dynamically after each step, HomeEnv evaluates execution validity turn-by-turn. To operationalize this, let $m_t = \frac{1}{K}\sum_{k=1}^{K}\phi_k(s_t)$ denote the completion rate of blueprint conditions at step $t$. We formulate a step-wise reward function that balances incremental task advancement, overall success, and execution safety:
\begin{equation}
r_t^{(e)}(s_t, a_t) = \lambda_{\mathrm{prog}}(m_t - m_{t-1}) + \lambda_{\mathrm{succ}} \mathbf{1}\left[ t=T \land m_T = 1 \right] - c_{\text{audit}} \mathbf{1}\left[ \text{blocked}(a_t) \right]
\end{equation}
Here, the progress term provides dense positive guidance whenever an action advances the household state closer to the overall goal. The success term strictly rewards full blueprint fulfillment at the episode's end. Meanwhile, the final indicator triggers a negative penalty $c_{\text{audit}}$ for illegal actions intercepted by the HomeEnv firewall. This composite structure effectively mitigates the sparse reward problem in long-horizon interactions while tightly bounding the agent's trial-and-error behavior within safe operational limits.


\section{Experiments}

\subsection{Experimental Setup}

\textbf{Datasets.} To evaluate our method, we constructed SmartHome-Bench based on HomeEnv. Comprising a total of 1,678 instances, the dataset comprehensively covers diverse room layouts and rich device compositions. Specifically, the benchmark evaluates five core tasks: Atomic Control (QT1) involves executing explicit and straightforward user commands; Compositional Control (QT2) requires orchestrating multiple devices and performing condition-dependent reasoning based on the environment; Ambiguous Intent (QT3) focuses on inferring or clarifying specific operational goals from vague user descriptions; Context-Aware Multi-turn Interaction (QT4) handles coreference resolution, topic switching, and backtracking across consecutive dialogue turns; and Personalized Memory (QT5) assesses the model's ability to adhere to user habits and preferences. More details regarding the dataset are provided in Appendix~\ref{app:dataset}.

\textbf{Baselines.} We compare our method against two categories of baselines. The first category comprises state-of-the-art LLMs, such as GPT-5.5~\cite{openai2026gpt55}, Claude-4.6 Sonnet~\cite{anthropic2026claude46sonnet}, Qwen3.5 Plus~\cite{qwen2026qwen35plus}, GLM-5~\cite{zeng2026glm5}, DeepSeek-V4-Flash~\cite{deepseek2026v4} and MiniMax M2.7~\cite{minimax2026m27}. For the trained baselines, all models are based on the Qwen3-4B and Qwen3-8B. The primary comparison baseline employs the conventional role-play data synthesis method~\cite{chen2024roleplaysurvey,abdullin2024syntheticdialogue}, which fine-tunes the base models using purely LLM-simulated trajectories. To ensure a fair comparison, the role-play baseline is trained with the same number of samples as our method.

\textbf{Evaluation Metrics.} We evaluate the success rate for each category and the overall rate via objective state verification within HomeEnv. Additionally, we assess operational efficiency by tracking the average number of tool calls and output tokens per task.

\subsection{Main Results}


\begin{table}[t]
  \centering
  \setlength{\tabcolsep}{6pt} 
  \caption{Performance (\%) of various LLMs on SmartHome-Bench under a unified agent framework.}
  \label{tab:model_performance}
  \scalebox{0.9}{ 
  \begin{tabular}{lcccccccc}
    \toprule
    \textbf{Model} & \textbf{QT1} & \textbf{QT2} & \textbf{QT3} & \textbf{QT4} & \textbf{QT5} & \textbf{Overall} & \begin{tabular}[c]{@{}c@{}}\textbf{Avg.}\\ \textbf{Tool Calls}\end{tabular} & \begin{tabular}[c]{@{}c@{}}\textbf{Avg.}\\ \textbf{Tokens (K)}\end{tabular} \\
    \midrule
    GPT-5.5 & 72.82 & \textbf{90.19} & \textbf{94.27} & 88.05 & 78.95 & \textbf{85.80} & 2.13 & 61.16\\
    Claude-4.6-Sonnet & 88.65 & 79.26 & 88.53 & 83.90 & 81.05 & 84.09 & 2.16 & 98.09\\
    Qwen-3.5 Plus & 89.45 & 73.89 & 88.53 & \textbf{89.87} & 85.26 & 84.15 & 2.17 & 84.03\\
    GLM-5 & 91.03 & 74.07 & 90.68 & 87.27 & \textbf{87.37} & 84.40 & 2.32 & 86.06\\
    DeepSeek-V4-Flash & \textbf{93.14} & 78.15 & 84.59 & 87.79 & 84.21 & 85.16 & 2.50 & 26.98\\
    MiniMax M2.7 & 73.88 & 55.56 & 61.65 & 81.30 & 81.05 & 68.06 & 2.10 & 77.90\\
    \midrule
    Qwen3-4B & 38.00 & 9.44 & 13.62 & 65.45 & 53.68 & 30.85 & 1.51 & 68.43 \\
    Role-play-4B & 89.71 & 55.56 & 82.08 & 85.45 & 74.74 & 75.62 & 1.96 & 119.01 \\
    HomeFlow-SFT-4B & 94.99 & 64.70 & \textbf{89.50} & 86.75 & 76.80 & 81.50 & 2.06 & 90.48 \\
    HomeFlow-RL-4B & \textbf{96.22} & \textbf{72.22} & 88.17 & \textbf{92.83} & \textbf{76.84} & \textbf{84.60} & 2.06 & 89.07 \\
    \midrule
    Qwen3-8B & 27.70 & 10.19 & 13.26 & 68.83 & 60.00 & 31.87 & 1.43 & 64.32 \\
    Role-play-8B & 91.56 & 60.19 & 84.95 & 85.45 & 70.53 & 77.78 & 1.97 & 119.65 \\
    HomeFlow-SFT-8B & \textbf{94.99} & 73.89 & \textbf{90.68} & 85.71 & 81.05 & 82.45 & 2.01 & 113.88 \\
    HomeFlow-RL-8B & 94.72 & \textbf{79.63} & \textbf{90.68} & \textbf{88.31} & \textbf{82.11} & \textbf{87.03} & 2.07 & 89.11 \\
    \bottomrule
  \end{tabular}
  }
\end{table}

Table~\ref{tab:model_performance} compares the performance of HomeFlow against various baselines on SmartHome-Bench. We observe that HomeFlow-RL, our full training pipeline, achieves the highest success rates across both Qwen3-4B and Qwen3-8B backbones. Specifically, for Qwen3-8B, HomeFlow-SFT achieves an overall success rate of 82.45\%, outperforming the Role-play by 4.67 points using the same number of training samples. Further optimization via step-wise RLVE elevates the performance to 87.03\%. A consistent trend is observed at the 4B scale, where HomeFlow-RL reaches a success rate of 84.60\%, significantly surpassing its base and role-play counterparts.

The results highlight three key advantages of our method. First, our verifiable data synthesis provides a superior cold start by directly mapping user instructions to verifiable physical state transitions, allowing the model to master complex device-control logic more reliably than pure-text simulation. Second, Step-wise RLVE ensures the agent's policy remains robust during dynamic, multi-turn interactions, enabling it to learn from real-time feedback and recover from compounding errors that typically lead to task failure. Finally, our 8B HomeFlow-RL exceeds the performance of leading proprietary LLMs, proving that training within an interactive execution environment can overcome model scale constraints when handling complex, multi-device household orchestrations.

\begin{table}[t]
  \centering
  \footnotesize
  \setlength{\tabcolsep}{10pt}
  \caption{Analysis of asymmetric node expansion budgets. Base Traj. denotes the initial linear trajectories, Explored Traj. represents the additional trajectories yielded by node expansion in MCTS-Flow, and Total Traj. is their combined volume.}
  \label{tab:node_expansion_ablation}
  \begin{tabular}{cccccc}
    \toprule
    \textbf{$k_{\text{user}}$} & \textbf{$k_{\text{agent}}$} & \textbf{Base Traj.} & \textbf{Explored Traj.} & \textbf{Total Traj.} & \textbf{Success Rate (\%)} \\
    \midrule
    1 & 1 & 22,000 & 0     & 22,000 & 80.4 \\
    3 & 1 & 15,000 & 5,662 & 20,662 & 82.6 \\
    1 & 3 & 15,000 & 6,708 & 21,708 & 84.7 \\
    3 & 3 & 15,000 & 6,871 & 21,871 & 85.3 \\
    \bottomrule
  \end{tabular}
\end{table}

\subsection{Analysis of Data Synthesis}

\subsubsection{Comparison of Synthesis Paradigms}

\begin{wrapfigure}{r}{0.5\textwidth}
  \centering
  \includegraphics[width=\linewidth]{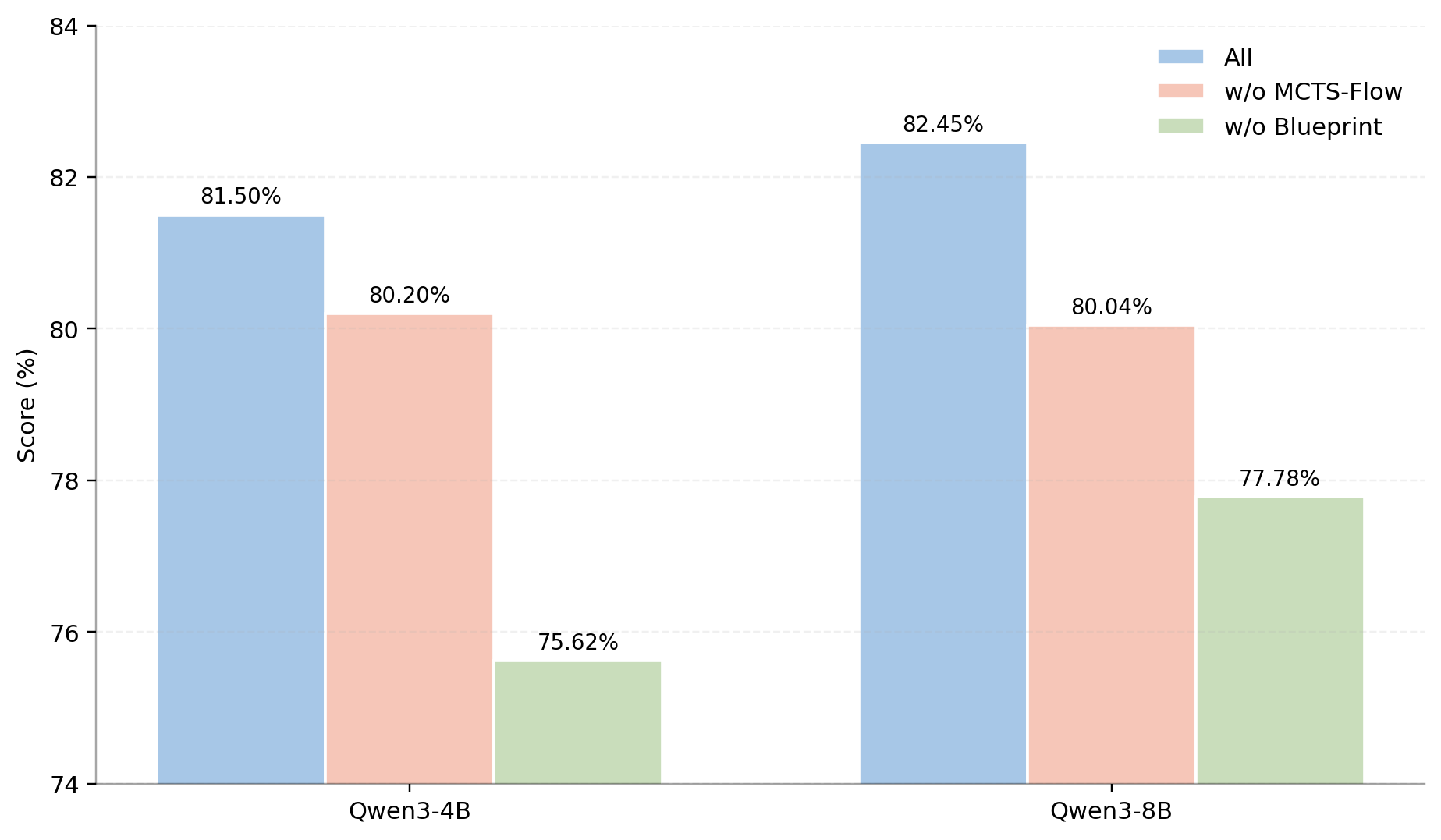}
  \captionof{figure}{Performance comparison of different SFT data synthesis strategies. Our approach is evaluated against linear generation (w/o MCTS-Flow) and pure-text role-play (w/o Blueprint) to demonstrate the necessity of each component.}
  \label{fig:ablation_sft}
\end{wrapfigure}

We trace the performance improvements during the SFT stage by progressively ablating the components of our data synthesis pipeline, as illustrated in Figure \ref{fig:ablation_sft}. Our complete approach employs Blueprint grounding coupled with MCTS-Flow exploration. Removing the MCTS-Flow component (w/o MCTS-Flow) degrades the pipeline to Blueprint-grounded linear generation. Further removing the Blueprint constraints (w/o Blueprint) reduces the method to conventional pure-text role-play. This comparison allows us to assess both the efficiency of the data generation process and its impact on downstream model performance.

\textbf{Higher Data Synthesis Efficiency.} We first evaluate the generation efficiency of these paradigms as an empirical phenomenon. Pure-text role-play suffers from LLM hallucinations, yielding a dialogue generation success rate of 86.33\%. Introducing physically grounded Blueprints forces the generation to respect initial states and tool logic, raising the success rate to 92.43\%. However, this linear generation still discards the entire dialogue if a single step fails. By incorporating tree-based exploration and a rollout-saving mechanism, the full MCTS-Flow pipeline avoids this waste, successfully completing complex trajectories and achieving the highest generation success rate of 96.48\%.

\textbf{Verifiable Trajectories Provide a Better Cold Start.} The quality of this synthesized data directly translates to downstream model performance. Upgrading from pure-text role-play to Blueprint-grounded linear generation raises the task success rate on Qwen3-4B and Qwen3-8B by 4.58 and 2.26 percentage points, respectively. This demonstrates that trajectories generated via pure-text role-play, while textually plausible, often violate hidden device preconditions. In contrast, Blueprint-grounded trajectories are executed and verified in the HomeEnv sandbox, enabling the model to learn executable state transitions rather than mimicking surface-level conversational patterns.

\textbf{MCTS Enhances Trajectory Diversity.} Building upon the Blueprint-grounded generation, incorporating MCTS-Flow exploration further improves SFT performance, reaching a task success rate of 82.45\% on Qwen3-8B and 81.50\% on Qwen3-4B. By searching through alternative user intents and tool invocation sequences, MCTS-Flow significantly increases the diversity and coverage of successful trajectories. This mechanism proves particularly effective for complex, multi-step scenarios, establishing a robust behavioral prior before the multi-turn RL optimization.

\subsubsection{Impact of Asymmetric Branching Factors in MCTS-Flow}

Smart home interactions exhibit an inherent role-based asymmetry: user utterances are largely deterministic and constrained by blueprints, whereas agent decisions involve complex multi-step reasoning. To analyze how this affects search efficiency, we evaluate different combinations of branching factors $k_{\text{user}}$ and $k_{\text{agent}}$ while maintaining a comparable total trajectory scale, as shown in Table \ref{tab:node_expansion_ablation}. We observe that allocating a larger budget to agent nodes where $k_{\text{agent}}$ equals 3 yields a significantly higher success rate of 84.7\% compared to the 82.6\% achieved by user-side expansion. This confirms that the agent’s action space is the primary source of task complexity. While joint expansion where both $k_{\text{user}}$ and $k_{\text{agent}}$ equal 3 achieves the highest success rate of 85.3\%, the marginal improvement over agent-only expansion is small. These results empirically justify our asymmetric branching strategy where $k_{\text{agent}} > k_{\text{user}}$, which effectively concentrates the search budget on critical agent decision bifurcations while avoiding redundant sampling of deterministic user inputs.


\begin{wrapfigure}{r}{0.5\textwidth}
  \centering
  
  \footnotesize
  \setlength{\tabcolsep}{4pt}
  \captionof{table}{Comparison of performance (\%) across different training paradigms. \textbf{ST} denotes single-turn RL where updates only occur at the final step, while \textbf{Full} refers to our complete Step-wise RLVE with multi-turn episodic rewards.}
  \label{tab:component_ablation_refined}
  
  \begin{tabular}{lccc}
    \toprule
    \textbf{Model} & \begin{tabular}[c]{@{}c@{}}\textbf{Success}\\ \textbf{Rate}\end{tabular} & \begin{tabular}[c]{@{}c@{}}\textbf{Avg.}\\ \textbf{Tools}\end{tabular} & \begin{tabular}[c]{@{}c@{}}\textbf{Avg.}\\ \textbf{Toks}\end{tabular} \\
    \midrule
    HomeFlow-SFT       & 82.45          & 2.01 & 113.8 \\
    HomeFlow-RL (ST)   & 86.46          & 2.25 & 79.0  \\
    HomeFlow-RL (Full) & \textbf{87.03} & 2.07 & 89.1  \\
    \bottomrule
  \end{tabular}
  
  \vspace{1.5em} 
  
  \includegraphics[width=\linewidth]{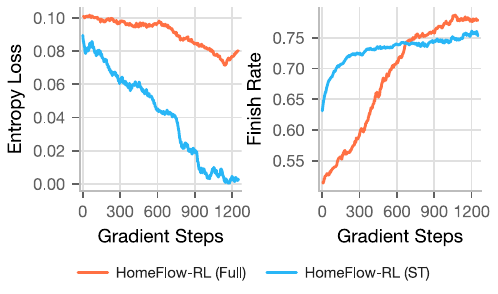}
  \captionof{figure}{Training dynamics of HomeFlow-RL variants. Sparse feedback causes early policy collapse in the ST variant, whereas dense step-wise rewards in the complete model sustain exploration.}
  \label{fig:rl-single-vs-multi}
\end{wrapfigure}

\subsection{Analysis of Step-wise RLVE}

Step-wise RLVE serves as a critical phase to align the agent's policy with the non-stationary dynamics of physical households. While MCTS-enhanced SFT provides a high-quality behavioral prior, Step-wise RLVE enables the agent to transition from passive imitation to environment-driven exploration. By interacting with a dynamic user simulator and the HomeEnv sandbox, the agent learns to refine its decision-making through trial-and-error, grounded by composite verifiable rewards. As shown in Table~\ref{tab:component_ablation_refined}, the HomeFlow-RL pipeline achieves a task success rate of 87.03\%, representing a substantial 4.58 percentage point improvement over the HomeFlow-SFT baseline. This gain demonstrates that episodic optimization allows the agent to manage long-horizon dependencies and recover from intermediate execution failures, effectively overcoming the compounding errors that typically plague static imitation learning.

To further investigate reward granularity, we compare HomeFlow-RL with HomeFlow-RL (ST). In the latter, the agent perceives the entire context but updates are restricted to the final action based on the ultimate outcome. While this improves success rates to 86.46\%, the scarcity of feedback makes the policy vulnerable to early collapse. As shown in Figure~\ref{fig:rl-single-vs-multi}, the policy entropy of HomeFlow-RL (ST) drops sharply, whereas HomeFlow-RL maintains exploration through dense, real-time guidance. Notably, HomeFlow-RL reduces average tool calls by 8\% compared to this variant, suggesting that step-wise credit assignment fosters more concise and purposeful planning. These results justify the necessity of dense rewards in verifiable environments. Qualitative examples of post-training model execution are provided in Appendix~\ref{app:case_study}.

\section{Conclusion}

HomeFlow establishes a verifiable training pipeline where simulation specifies tasks, verifies trajectories, and provides rule-based RL rewards. Combining Blueprint-grounded synthesis, MCTS expansion, and step-wise RLVE, HomeFlow boosts the Qwen-4B and Qwen-8B models to 84.20\% and 87.03\% task success rates on SmartHome-Bench, respectively. Our results yield three key insights: (1) verifiable environments provide a unified foundation for both SFT data synthesis and online policy optimization; (2) step-wise interaction sustains exploration and significantly improves long-horizon control; and (3) MCTS-driven search enhances trajectory diversity while transforming verified prefixes into reusable assets. Ultimately, these findings suggest a practical recipe for AIoT agents with checkable goals.


\textbf{Limitations and future work.} Currently, HomeFlow relies on deterministic simulation, necessitating sim-to-real adaptation to address real-world uncertainties like latency and sensor drift. Broader deployment also requires adapting to diverse vendor-specific protocols to expand device coverage, alongside mitigating the inherent computational overhead introduced by MCTS and online RL. Promising future directions include incorporating continual learning for personalized household routines and extending the verifiable framework to broader AIoT ecosystems.



\bibliography{references}

\begin{thebibliography}{43}
\providecommand{\natexlab}[1]{#1}

\bibitem[{Abdullin et~al.(2024)Abdullin, Molla-Aliod, Ofoghi, Yearwood, and Li}]{abdullin2024syntheticdialogue}
Yelaman Abdullin, Diego Molla-Aliod, Bahadorreza Ofoghi, John Yearwood, and Qingyang Li. 2024.
\newblock \href {https://arxiv.org/abs/2401.17461} {Synthetic dialogue dataset generation using llm agents}.
\newblock \emph{arXiv preprint arXiv:2401.17461}.

\bibitem[{{Anthropic}(2026)}]{anthropic2026claude46sonnet}
{Anthropic}. 2026.
\newblock \href {https://www.anthropic.com/claude-sonnet-4-6-system-card} {Claude sonnet 4.6 system card}.
\newblock \url{https://www.anthropic.com/claude-sonnet-4-6-system-card}.
\newblock Accessed: 2026-05-07.

\bibitem[{Bai et~al.(2022)Bai, Kadavath, Kundu, Askell, Kernion, Jones, Chen, Goldie, Mirhoseini, McKinnon et~al.}]{bai2022constitutionalai}
Yuntao Bai, Saurav Kadavath, Sandipan Kundu, Amanda Askell, Jackson Kernion, Andy Jones, Anna Chen, Anna Goldie, Azalia Mirhoseini, Cameron McKinnon, and 1 others. 2022.
\newblock \href {https://arxiv.org/abs/2212.08073} {Constitutional ai: Harmlessness from ai feedback}.
\newblock \emph{arXiv preprint arXiv:2212.08073}.

\bibitem[{Birkmose et~al.(2025)Birkmose, Reece, Norvin, Bjerva, and Zhang}]{birkmose2025homeassistantllm}
Rune Birkmose, Nathan~M{\o}rkeberg Reece, Esben~Hofstedt Norvin, Johannes Bjerva, and Mike Zhang. 2025.
\newblock \href {https://arxiv.org/abs/2502.12923} {On-device llms for home assistant: Dual role in intent detection and response generation}.
\newblock \emph{arXiv preprint arXiv:2502.12923}.

\bibitem[{Chen et~al.(2024)Chen, Wang, Deng, and Li}]{chen2024roleplaysurvey}
Nuo Chen, Yan Wang, Yang Deng, and Jia Li. 2024.
\newblock \href {https://arxiv.org/abs/2407.11484} {The oscars of ai theater: A survey on role-playing with language models}.
\newblock \emph{arXiv preprint arXiv:2407.11484}.

\bibitem[{Comanici et~al.(2025)Comanici, Bieber, Schaekermann, Pasupat, Sachdeva, Dhillon, Blistein, Ram, Zhang, Rosen et~al.}]{comanici2025gemini25}
Gheorghe Comanici, Eric Bieber, Mike Schaekermann, Ice Pasupat, Noveen Sachdeva, Inderjit Dhillon, Marcel Blistein, Ori Ram, Dan Zhang, Evan Rosen, and 1 others. 2025.
\newblock \href {https://arxiv.org/abs/2507.06261} {Gemini 2.5: Pushing the frontier with advanced reasoning, multimodality, long context, and next generation agentic capabilities}.
\newblock \emph{arXiv preprint arXiv:2507.06261}.

\bibitem[{{DeepSeek-AI}(2026)}]{deepseek2026v4}
{DeepSeek-AI}. 2026.
\newblock \href {https://huggingface.co/deepseek-ai/DeepSeek-V4-Pro/blob/main/DeepSeek_V4.pdf} {Deepseek v4 technical documentation}.
\newblock \url{https://huggingface.co/deepseek-ai/DeepSeek-V4-Pro/blob/main/DeepSeek_V4.pdf}.
\newblock Accessed: 2026-05-07.

\bibitem[{Gao et~al.(2025)Gao, Ye, Wang, and Sang}]{gao2025websynthesis}
Yifei Gao, Junhong Ye, Jiaqi Wang, and Jitao Sang. 2025.
\newblock \href {https://arxiv.org/abs/2507.04370} {Websynthesis: World-model-guided mcts for efficient webui-trajectory synthesis}.
\newblock \emph{arXiv preprint arXiv:2507.04370}.

\bibitem[{Gu et~al.(2024)Gu, Jiang, Shi, Tan, Zhai, Xu, Li, Shen, Ma, Liu et~al.}]{gu2024llmjudge}
Jiawei Gu, Xuhui Jiang, Zhichao Shi, Hexiang Tan, Xuehao Zhai, Chengjin Xu, Wei Li, Yinghan Shen, Shengjie Ma, Honghao Liu, and 1 others. 2024.
\newblock \href {https://arxiv.org/abs/2411.15594} {A survey on llm-as-a-judge}.
\newblock \emph{arXiv preprint arXiv:2411.15594}.

\bibitem[{Guo et~al.(2025)}]{guo2025deepseekr1}
Daya Guo and 1 others. 2025.
\newblock \href {https://doi.org/10.1038/s41586-025-09422-z} {Deepseek-r1 incentivizes reasoning in llms through reinforcement learning}.
\newblock \emph{Nature}, 645:633--638.

\bibitem[{Hammi et~al.(2022)Hammi, Zeadally, Khatoun, and Nebhen}]{hammi2022smarthome}
Badis Hammi, Sherali Zeadally, Rida Khatoun, and Jamel Nebhen. 2022.
\newblock \href {https://doi.org/10.1016/j.cose.2022.102677} {Survey on smart homes: Vulnerabilities, risks, and countermeasures}.
\newblock \emph{Computers \& Security}, 117:102677.

\bibitem[{Huang et~al.(2025)Huang, Shen, Ma, and Zheng}]{huang2025homellama}
Xinyu Huang, Leming Shen, Zijing Ma, and Yuanqing Zheng. 2025.
\newblock \href {https://arxiv.org/abs/2507.08878} {Towards privacy-preserving and personalized smart homes via tailored small language models}.
\newblock \emph{arXiv preprint arXiv:2507.08878}.

\bibitem[{Huda et~al.(2024)Huda, Ahmed, Adnan, Ali, and Naeem}]{huda2024smarthomecities}
Noor~Ul Huda, Ijaz Ahmed, Muhammad Adnan, Mansoor Ali, and Faisal Naeem. 2024.
\newblock \href {https://doi.org/10.1016/j.eswa.2023.122380} {Experts and intelligent systems for smart homes' transformation to sustainable smart cities: A comprehensive review}.
\newblock \emph{Expert Systems with Applications}, 238:122380.

\bibitem[{Jimenez et~al.(2023)Jimenez, Yang, Wettig, Yao, Pei, Press, and Narasimhan}]{jimenez2023swebench}
Carlos~E. Jimenez, John Yang, Alexander Wettig, Shunyu Yao, Kexin Pei, Ofir Press, and Karthik Narasimhan. 2023.
\newblock \href {https://arxiv.org/abs/2310.06770} {Swe-bench: Can language models resolve real-world github issues?}
\newblock \emph{arXiv preprint arXiv:2310.06770}.

\bibitem[{King et~al.(2024)King, Yu, Lee, and Julien}]{king2024sasha}
Evan King, Haoxiang Yu, Sangsu Lee, and Christine Julien. 2024.
\newblock \href {https://doi.org/10.1145/3643505} {Sasha: Creative goal-oriented reasoning in smart homes with large language models}.
\newblock \emph{Proceedings of the ACM on Interactive, Mobile, Wearable and Ubiquitous Technologies}, 8(1):1--38.

\bibitem[{Koh et~al.(2024)Koh, Lo, Jang, Duvvur, Lim, Huang, Neubig, Fried, and Salakhutdinov}]{koh2024treesearchagents}
Jing~Yu Koh, Robert Lo, Lawrence Jang, Vinay Duvvur, Ming~Chong Lim, Po-Yao Huang, Graham Neubig, Daniel Fried, and Ruslan Salakhutdinov. 2024.
\newblock \href {https://arxiv.org/abs/2407.01476} {Tree search for language model agents}.
\newblock \emph{arXiv preprint arXiv:2407.01476}.

\bibitem[{Lee et~al.(2024)Lee, Phatale, Mansoor, Mesnard, Ferret, Lu, Bishop, Hall, Carbune, Rastogi et~al.}]{lee2024rlaif}
Harrison Lee, Samrat Phatale, Hassan Mansoor, Thomas Mesnard, Johan Ferret, Kellie Lu, Colton Bishop, Ethan Hall, Victor Carbune, Abhinav Rastogi, and 1 others. 2024.
\newblock \href {https://openreview.net/forum?id=AAxIs3D2ZZ} {Rlaif vs. rlhf: Scaling reinforcement learning from human feedback with ai feedback}.
\newblock In \emph{Proceedings of the 41st International Conference on Machine Learning}.

\bibitem[{Li et~al.(2024)Li, Dong, Chen, Su, Zhou, Ai, Ye, and Liu}]{li2024llmjudge}
Haitao Li, Qian Dong, Junjie Chen, Huixue Su, Yujia Zhou, Qingyao Ai, Ziyi Ye, and Yiqun Liu. 2024.
\newblock \href {https://arxiv.org/abs/2412.05579} {Llms-as-judges: A comprehensive survey on llm-based evaluation methods}.
\newblock \emph{arXiv preprint arXiv:2412.05579}.

\bibitem[{Li et~al.(2025{\natexlab{a}})Li, Chen, Qu, Xu, Lin, Zhu, Xu, Tan, Fu, Ju et~al.}]{li2025mimovlmiloco}
Jiaze Li, Jingyang Chen, Yuxun Qu, Shijie Xu, Zhenru Lin, Junyou Zhu, Boshen Xu, Wenhui Tan, Pei Fu, Jianzhong Ju, and 1 others. 2025{\natexlab{a}}.
\newblock \href {https://arxiv.org/abs/2512.17436} {Xiaomi mimo-vl-miloco technical report}.
\newblock \emph{arXiv preprint arXiv:2512.17436}.

\bibitem[{Li et~al.(2025{\natexlab{b}})Li, Guo, Yao, Liu, and Wang}]{li2025homebench}
Silin Li, Yuhang Guo, Jiashu Yao, Zeming Liu, and Haifeng Wang. 2025{\natexlab{b}}.
\newblock \href {https://arxiv.org/abs/2505.19628} {Homebench: Evaluating llms in smart homes with valid and invalid instructions across single and multiple devices}.
\newblock \emph{arXiv preprint arXiv:2505.19628}.

\bibitem[{{MiniMax}(2026)}]{minimax2026m27}
{MiniMax}. 2026.
\newblock \href {https://www.minimax.io/news/minimax-m27-en} {Minimax m2.7: Early echoes of self-evolution}.
\newblock \url{https://www.minimax.io/news/minimax-m27-en}.
\newblock Accessed: 2026-05-07.

\bibitem[{Mitra et~al.(2024)Mitra, Del~Corro, Zheng, Mahajan, Rouhana, Codas, Lu, Chen, Vrousgos, Rosset et~al.}]{mitra2024agentinstruct}
Arindam Mitra, Luciano Del~Corro, Guoqing Zheng, Shweti Mahajan, Dany Rouhana, Andres Codas, Yadong Lu, Wei-ge Chen, Olga Vrousgos, Corby Rosset, and 1 others. 2024.
\newblock \href {https://arxiv.org/abs/2407.03502} {Agentinstruct: Toward generative teaching with agentic flows}.
\newblock \emph{arXiv preprint arXiv:2407.03502}.

\bibitem[{Ni et~al.(2026)Ni, Wang, Wang, Kveton, Dernoncourt, Xia, Chen, Leura, Basu, Mukherjee et~al.}]{ni2026usersimulation}
Bo~Ni, Leyao Wang, Yu~Wang, Branislav Kveton, Franck Dernoncourt, Yu~Xia, Hongjie Chen, Reuben Leura, Samyadeep Basu, Subhojyoti Mukherjee, and 1 others. 2026.
\newblock \href {https://arxiv.org/abs/2604.24977} {A survey on llm-based conversational user simulation}.
\newblock \emph{arXiv preprint arXiv:2604.24977}.

\bibitem[{{OpenAI}(2026)}]{openai2026gpt55}
{OpenAI}. 2026.
\newblock \href {https://deploymentsafety.openai.com/gpt-5-5} {Gpt-5.5 system card}.
\newblock \url{https://deploymentsafety.openai.com/gpt-5-5}.
\newblock Accessed: 2026-05-07.

\bibitem[{Ouyang et~al.(2022)Ouyang, Wu, Jiang, Almeida, Wainwright, Mishkin, Zhang, Agarwal, Slama, Ray et~al.}]{ouyang2022instructgpt}
Long Ouyang, Jeff Wu, Xu~Jiang, Diogo Almeida, Carroll~L. Wainwright, Pamela Mishkin, Chong Zhang, Sandhini Agarwal, Katarina Slama, Alex Ray, and 1 others. 2022.
\newblock \href {https://proceedings.neurips.cc/paper_files/paper/2022/hash/b1efde53be364a73914f58805a001731-Abstract-Conference.html} {Training language models to follow instructions with human feedback}.
\newblock In \emph{Advances in Neural Information Processing Systems}.

\bibitem[{{Qwen Team}(2026)}]{qwen2026qwen35plus}
{Qwen Team}. 2026.
\newblock \href {https://qwen.ai/blog?id=qwen3.5} {Qwen3.5: Towards native multimodal agents}.
\newblock \url{https://qwen.ai/blog?id=qwen3.5}.
\newblock Accessed: 2026-05-07.

\bibitem[{Rivkin et~al.(2023)Rivkin, Hogan, Feriani, Konar, Sigal, Liu, and Dudek}]{rivkin2023sage}
Dmitriy Rivkin, Francois Hogan, Amal Feriani, Abhisek Konar, Adam Sigal, Steve Liu, and Greg Dudek. 2023.
\newblock \href {https://arxiv.org/abs/2311.00772} {Sage: Smart home agent with grounded execution}.
\newblock \emph{arXiv preprint arXiv:2311.00772}.

\bibitem[{Seo et~al.(2026)Seo, Yang, Pyo, Kim, Lee, and Jo}]{seo2026simuhome}
Gyuhyeon Seo, Jungwoo Yang, Junseong Pyo, Nalim Kim, Jonggeun Lee, and Yohan Jo. 2026.
\newblock \href {https://openreview.net/forum?id=LCS1WsGvha} {Simuhome: A temporal- and environment-aware benchmark for smart home llm agents}.
\newblock In \emph{International Conference on Learning Representations}.

\bibitem[{Shao et~al.(2023)Shao, Li, Dai, and Qiu}]{shao2023characterllm}
Yunfan Shao, Linyang Li, Junqi Dai, and Xipeng Qiu. 2023.
\newblock \href {https://aclanthology.org/2023.emnlp-main.814/} {Character-llm: A trainable agent for role-playing}.
\newblock In \emph{Proceedings of the 2023 Conference on Empirical Methods in Natural Language Processing}, pages 13153--13187.

\bibitem[{Shao et~al.(2024)Shao, Wang, Zhu, Xu, Song, Bi, Zhang, Zhang, Li, Wu, and Guo}]{shao2024deepseekmath}
Zhihong Shao, Peiyi Wang, Qihao Zhu, Runxin Xu, Junxiao Song, Xiao Bi, Haowei Zhang, Mingchuan Zhang, Y.~K. Li, Y.~Wu, and Daya Guo. 2024.
\newblock \href {https://arxiv.org/abs/2402.03300} {Deepseekmath: Pushing the limits of mathematical reasoning in open language models}.
\newblock \emph{arXiv preprint arXiv:2402.03300}.

\bibitem[{Singh et~al.(2025)Singh, Fry, Perelman, Tart, Ganesh, El-Kishky, McLaughlin, Low, Ostrow, Ananthram et~al.}]{singh2025gpt5}
Aaditya Singh, Adam Fry, Adam Perelman, Adam Tart, Adi Ganesh, Ahmed El-Kishky, Aidan McLaughlin, Aiden Low, AJ~Ostrow, Akhila Ananthram, and 1 others. 2025.
\newblock \href {https://arxiv.org/abs/2601.03267} {Openai gpt-5 system card}.
\newblock \emph{arXiv preprint arXiv:2601.03267}.

\bibitem[{Wang et~al.(2025)Wang, Wang, Zhang, Yuan, Xu, Huang, Yuan, Guo, Chen, Zhou et~al.}]{wang2025coser}
Xintao Wang, Heng Wang, Yifei Zhang, Xinfeng Yuan, Rui Xu, Jen-tse Huang, Siyu Yuan, Haoran Guo, Jiangjie Chen, Shuchang Zhou, and 1 others. 2025.
\newblock \href {https://proceedings.mlr.press/v267/wang25dk.html} {Coser: Coordinating llm-based persona simulation of established roles}.
\newblock In \emph{Proceedings of the Forty-Second International Conference on Machine Learning}.

\bibitem[{Wang et~al.(2022)Wang, Kordi, Mishra, Liu, Smith, Khashabi, and Hajishirzi}]{wang2022selfinstruct}
Yizhong Wang, Yeganeh Kordi, Swaroop Mishra, Alisa Liu, Noah~A. Smith, Daniel Khashabi, and Hannaneh Hajishirzi. 2022.
\newblock \href {https://arxiv.org/abs/2212.10560} {Self-instruct: Aligning language models with self-generated instructions}.
\newblock \emph{arXiv preprint arXiv:2212.10560}.

\bibitem[{Wei et~al.(2025)Wei, Yao, Liu, Zhang, Lu, Qiu, Yu, Xu, Zhang, Yin, Yun, and Li}]{wei2025webagentr1}
Zhepei Wei, Wenlin Yao, Yao Liu, Weizhi Zhang, Qin Lu, Liang Qiu, Changlong Yu, Puyang Xu, Chao Zhang, Bing Yin, Hyokun Yun, and Lihong Li. 2025.
\newblock \href {https://arxiv.org/abs/2505.16421} {Webagent-r1: Training web agents via end-to-end multi-turn reinforcement learning}.
\newblock \emph{arXiv preprint arXiv:2505.16421}.

\bibitem[{Xi et~al.(2025)Xi, Chen, Guo, He, Ding, Hong, Zhang, Wang, Jin, Zhou et~al.}]{xi2025llmagents}
Zhiheng Xi, Wenxiang Chen, Xin Guo, Wei He, Yiwen Ding, Boyang Hong, Ming Zhang, Junzhe Wang, Senjie Jin, Enyu Zhou, and 1 others. 2025.
\newblock \href {https://link.springer.com/article/10.1007/s11432-024-4222-0} {The rise and potential of large language model based agents: A survey}.
\newblock \emph{Science China Information Sciences}, 68(2):121101.

\bibitem[{Xu et~al.(2023)Xu, Sun, Zheng, Geng, Zhao, Feng, Tao, Lin, and Jiang}]{xu2023wizardlm}
Can Xu, Qingfeng Sun, Kai Zheng, Xiubo Geng, Pu~Zhao, Jiazhan Feng, Chongyang Tao, Qingwei Lin, and Daxin Jiang. 2023.
\newblock \href {https://arxiv.org/abs/2304.12244} {Wizardlm: Empowering large pre-trained language models to follow complex instructions}.
\newblock \emph{arXiv preprint arXiv:2304.12244}.

\bibitem[{Yang et~al.(2025)Yang, Li, Yang, Zhang, Hui, Zheng, Yu, Gao, Huang, Lv et~al.}]{yang2025qwen3}
An~Yang, Anfeng Li, Baosong Yang, Beichen Zhang, Binyuan Hui, Bo~Zheng, Bowen Yu, Chang Gao, Chengen Huang, Chenxu Lv, and 1 others. 2025.
\newblock \href {https://arxiv.org/abs/2505.09388} {Qwen3 technical report}.
\newblock \emph{arXiv preprint arXiv:2505.09388}.

\bibitem[{Yin et~al.(2025)Yin, Zhang, and Kawahara}]{yin2025harmony}
Ziqi Yin, Mingxin Zhang, and Daisuke Kawahara. 2025.
\newblock \href {https://doi.org/10.1145/3714394.3754356} {Harmony: A human-aware, responsive, modular assistant with a locally deployed large language model}.
\newblock In \emph{Companion of the 2025 ACM International Joint Conference on Pervasive and Ubiquitous Computing}, pages 126--130. Association for Computing Machinery.

\bibitem[{Yu et~al.(2026)Yu, Choi, Lee, Kim, Ko, Ko, and Oh}]{yu2026iotgpt}
Chaerin Yu, Chihun Choi, Sunjae Lee, Hyosu Kim, Steven~Y. Ko, Young-Bae Ko, and Sangeun Oh. 2026.
\newblock \href {https://arxiv.org/abs/2601.04680} {Leveraging llms for efficient and personalized smart home automation}.
\newblock \emph{arXiv preprint arXiv:2601.04680}.

\bibitem[{Yu et~al.(2025)Yu, Zhang, Zhu, Yuan, Zuo, Yue, Dai, Fan, Liu, Liu et~al.}]{yu2025dapo}
Qiying Yu, Zheng Zhang, Ruofei Zhu, Yufeng Yuan, Xiaochen Zuo, Yu~Yue, Weinan Dai, Tiantian Fan, Gaohong Liu, Lingjun Liu, and 1 others. 2025.
\newblock \href {https://arxiv.org/abs/2503.14476} {Dapo: An open-source llm reinforcement learning system at scale}.
\newblock \emph{arXiv preprint arXiv:2503.14476}.

\bibitem[{Zeng et~al.(2026)}]{zeng2026glm5}
Aohan Zeng and 1 others. 2026.
\newblock \href {https://arxiv.org/abs/2602.15763} {Glm-5: From vibe coding to agentic engineering}.
\newblock \emph{arXiv preprint arXiv:2602.15763}.

\bibitem[{Zhan et~al.(2026)Zhan, Li, Zhang, and Haddadi}]{zhan2026hearthnet}
Zhonghao Zhan, Krinos Li, Yefan Zhang, and Hamed Haddadi. 2026.
\newblock \href {https://arxiv.org/abs/2604.09618} {Hearthnet: Edge multi-agent orchestration for smart homes}.
\newblock \emph{arXiv preprint arXiv:2604.09618}.

\bibitem[{Zhang et~al.(2025)Zhang, Zuo, He, Sun, Liu, Jiang, Fan, Tian, Jia, Li et~al.}]{zhang2025rlreasoningsurvey}
Kaiyan Zhang, Yuxin Zuo, Bingxiang He, Youbang Sun, Runze Liu, Che Jiang, Yuchen Fan, Kai Tian, Guoli Jia, Pengfei Li, and 1 others. 2025.
\newblock \href {https://arxiv.org/abs/2509.08827} {A survey of reinforcement learning for large reasoning models}.
\newblock \emph{arXiv preprint arXiv:2509.08827}.

\end{thebibliography}

\newpage
\appendix
\clearpage

\section{HomeEnv Simulation Engine}
\label{app:homeenv}

We introduce \textsc{HomeEnv}, a smart home simulation engine that supports configurable device specifications, programmatic control APIs, and verifiable task execution in a Gym-style environment.

\subsection{YAML-based Smart Device Specification}

Devices are specified using a YAML-based configuration format. The following example defines a smart light device:

\begin{schemaformatbox}{YAML Specification of a Smart Light Device}
- name: light
  userdata:
    category: light
    brand: example-brand
    spid: "100001"
  attributes:
  - name: state
    type: str
    options: ["on", "off"]
  - name: brightness
    type: int
    range: [1, 100]
    unit: percentage
  - name: hs_color
    type: tuple
    items:
      - type: float
        range: [0.0, 360]
      - type: float
        range: [0, 100]
  services:
  - name: turn_on
    code: self.state = "on"
  - name: turn_off
    code: self.state = "off"
  - name: set_brightness
    arguments:
      - name: brightness
        type: int
        range: [1, 100]
    code: self.brightness = brightness
  - name: set_hs_color
    arguments:
      - name: hs_color
        type: tuple
        items:
          - type: float
            range: [0.0, 360]
          - type: float
            range: [0, 100]
    code: self.hs_color = hs_color
\end{schemaformatbox}

\subsection{Programmatic Device Interaction API}

Configured devices can be instantiated and controlled through a unified high-level API:

\begin{schemaformatbox}{Programmatic Control of a Configured Device}
# load configured light device
with open("light.yaml", "r") as f:
    entities = Entity.load(f)
    light_spec = entities[0]

# initialize device with randomized attributes
light = light_spec.rand()

# inspect and control device
print(light.state, light.brightness)

light.turn_on()
light.set_brightness(80)

print(light.state, light.brightness)
\end{schemaformatbox}

\subsection{Gym-style Home Simulation Environment}

We further build a Gym-style smart home environment that automatically generates house layouts and embedded devices, including rooms, floors, and device instances.

\begin{schemaformatbox}{Interaction with the HomeEnv Simulator}
env = gym.make("HomeEnv-v0", home=HomeSampler.sample_test())

observation, _, terminated, truncated, info = env.step({
    "name": "pyexec",
    "code": """
device("1001").set_brightness(80)
print(f"brightness of light is {device('1001').brightness}")
"""
})
\end{schemaformatbox}

\subsection{Verifiable Task Evaluation}

To support automatic evaluation, we define verifiable tasks as logical conditions derived from the environment state.

\begin{schemaformatbox}{Verification of Environment Tasks}
# ground-truth conditions
ground_truth_list = [
    "device('1001').state == 'on'",
    "device('1001').brightness > 60",
]

env = gym.make("HomeEnv-v0", home=HomeSampler.sample_test())

# register tasks
for gt in ground_truth_list:
    env.engine.task.add_task(gt)

# interact with environment
env.step({"name": "pyexec", "code": "..."})

# verify task completion
for task in env.engine.task.tasks:
    print(f"task id={task.id}  status={task.finished}  condition={task.condition}")
\end{schemaformatbox}
\section{Implementation Details of Blueprint}

We propose a two-stage data synthesis framework for generating large-scale, controllable, and diverse task-oriented dialogue data. The overall pipeline consists of:

\textbf{Blueprint Generation.}
constructing structured task blueprints conditioned on user profiles, household environments, and constraint validation mechanisms.

\textbf{Dialogue Generation.}
synthesizing grounded multi-turn interactions based on generated blueprints, user profiles, and simulated agents.

\subsection{Blueprint Generation}

\subsubsection{Meta Information Construction}

Blueprint generation is conditioned on three categories of meta-information: household environment, user profile, and tool/action space,detail information in Table~\ref{tab:meta_info}.

\begin{table}[!htbp]
\centering
\caption{Meta-information used in blueprint construction.}
\label{tab:meta_info}
\resizebox{\linewidth}{!}{%
\begin{tabular}{p{3.2cm} p{5.8cm} p{6cm}}
\toprule
\textbf{Category} & \textbf{Attributes} & \textbf{Description} \\
\midrule

Household Environment 
& temperature, humidity, brightness 
& Physical state variables in a smart-home environment. \\
& device APIs (e.g., light.turn\_on, ac.set\_temp) 
& Controllable smart home device interfaces. \\
& tool APIs (e.g., calculator, weather) 
& External function invocation interfaces. \\
& environment entities 
& Spatial context such as rooms and indoor/outdoor zones. \\

\midrule

User Profile 
& age, gender 
& Basic demographic attributes. \\
& physiological attributes 
& Height, weight, and physical condition. \\
& health status 
& Chronic diseases and sleep-related conditions. \\
& behavior habits 
& Daily routines and lifestyle patterns. \\
& preferences 
& Environmental preferences (temperature, lighting, etc.). \\

\bottomrule
\end{tabular}%
}
\end{table}

To ensure realism and diversity, we sample attributes from predefined distributions and enforce structured consistency constraints (e.g., age–health correlation, lifestyle–preference alignment).

We adopt a three-stage generation paradigm:

\textbf{(1) Base Blueprint Generator.}
Given user profiles and environment states, the generator produces a candidate set of task blueprints representing possible user goals.The overall Blueprint as shown in Section~\ref{sec:full-example}.

\textbf{(2) Blueprint Filter.}
The Blueprint Filter removes invalid or low-quality candidates before task execution,
as summarized in Table~\ref{tab:blueprint-filter}.

\begin{table}[htbp]
\centering
\caption{Filtering criteria used by the Blueprint Filter.}
\vspace{0.5em}
\begin{tabular}{ll}
\toprule
\textbf{Criterion} & \textbf{Description} \\
\midrule
Logical consistency & Removes contradictory or incoherent candidates. \\
Executability & Ensures compatibility with environment and tool APIs. \\
Diversity & Reduces redundant candidates and improves task coverage. \\
\bottomrule
\end{tabular}
\label{tab:blueprint-filter}
\end{table}

\textbf{(3) Query Type.}
We introduce a structured query perturbation mechanism to control difficulty and diversity,as summarized in Table~\ref{tab:query_control}.
\begin{table}[!htbp]
\centering
\caption{Query perturbation control space.}
\label{tab:query_control}
\resizebox{\linewidth}{!}{%
\begin{tabular}{p{3.2cm} p{4.2cm} p{6.5cm}}
\toprule
\textbf{Dimension} & \textbf{Options} & \textbf{Description} \\
\midrule

Intent Clarity 
& clear / ambiguous / missing-slot 
& Controls explicitness of user intent. \\

Noise Level 
& clean / noisy 
& Controls linguistic perturbation and ambiguity. \\

Device Scope 
& single-device / multi-device / batch-device& Controls action granularity. \\

Execution Dependency 
& independent / dependent 
& Controls temporal dependency across steps. \\

\bottomrule
\end{tabular}%
}
\end{table}

Formally, queries are sampled from the joint space of these factors to enable controlled complexity and diversity.

\subsubsection{Fully Example}
\label{sec:full-example}
\begin{tcolorbox}[
    breakable,
    colback=yellow!6!white,
    colframe=blue!50!green,
    title={\textbf{Blueprint}},
    sharp corners,
    boxrule=0.6pt,
    width=\linewidth
]
\small
\raggedright

\textbf{Mission:} Study reading and pre-exercise environment adjustment

\textbf{Description:}

Scene 1: In the afternoon, Li Mei prepares to practice handwriting in the study. Indoor sensors show the temperature is about 13\,°C; she feels a bit cold. The ceiling fan light is on but only at 11\% brightness with a slightly warm color temperature, which makes strokes hard to see. The smart speaker is paused but not muted, and she worries notification sounds may interrupt. She hopes to warm the room quickly, then make the lighting brighter and cooler white, and finally keep the environment quiet for focused writing.

Scene 2: In the spare bedroom doing tai chi warm-up, the hygrometer shows the room is hot (about 31.8\,°C) with relatively high humidity (about 62\%). He wants to lower the humidity and bring perceived temperature closer to around 20\,°C for light exercise.

\vspace{0.5em}
\textbf{Tasks:}

\begin{enumerate}
\item
\textbf{Intent:} The room feels cold—she wants it warmer; before writing, the light is too dim and too cool-toned—she wants it brighter.

\textbf{Query Type:} missing\_slot+quiet+dependent+multiple\_devices

\noindent\fcolorbox{red}{yellow!6!white}{%
\begin{minipage}{\dimexpr\linewidth-2\fboxsep-2\fboxrule\relax}
\textbf{Query:} If the AC is off, turn on heating and set the temperature to around 24\,°C; also the study light is too dim—raise both brightness and color temperature.
\end{minipage}}
\par\noindent
\begin{tcolorbox}[
    colframe=red,
    boxrule=0.85pt,
    colback=yellow!6!white,
    sharp corners,
    arc=0pt,
    boxsep=3pt,
    left=4pt,
    right=4pt,
    width=\linewidth
]
\textbf{Condition:}
\begin{lstlisting}[
language=Python,
columns=fullflexible,
keepspaces=true,
breaklines=true,
breakatwhitespace=false,
showstringspaces=false,
xleftmargin=4pt
]
(device("FQVZSo_p").state == "on" and 
 device("FQVZSo_p").ac_mode == "heat" and 
 23.5 <= device("FQVZSo_p").target_temperature <= 24.5) and 
(85 <= device("sTPi5CsC").light.brightness <= 100) and 
(5000 <= device("sTPi5CsC").light.color_temperature <= 6500)
\end{lstlisting}
\end{tcolorbox}

\textbf{Devices:} FQVZSo\_p, sTPi5CsC

\vspace{0.6em}

\item
\textbf{Intent:} Worried about being disturbed by notification sounds—wants a quieter environment.

\textbf{Query Type:} clear+quiet+dependent+single\_device

\noindent\fcolorbox{red}{yellow!6!white}{%
\begin{minipage}{\dimexpr\linewidth-2\fboxsep-2\fboxrule\relax}
\textbf{Query:} If the smart speaker is not muted, mute it.
\end{minipage}}
\par\noindent
\begin{tcolorbox}[
    colframe=red,
    boxrule=0.85pt,
    colback=yellow!6!white,
    sharp corners,
    arc=0pt,
    boxsep=3pt,
    left=4pt,
    right=4pt,
    width=\linewidth
]
\textbf{Condition:}
\begin{lstlisting}[
language=Python,
columns=fullflexible,
keepspaces=true,
breaklines=true,
breakatwhitespace=false,
showstringspaces=false,
xleftmargin=4pt
]
(device("1TirJUuR").player.volume_muted_enabled == True)
\end{lstlisting}
\end{tcolorbox}

\textbf{Devices:} 1TirJUuR

\vspace{0.6em}

\item
\textbf{Intent:} The spare bedroom humidity is high—wants it lower; the spare bedroom temperature is high—wants it cooler.

\textbf{Query Type:} clear+quiet+independent+sigle\_devices

\noindent\fcolorbox{red}{yellow!6!white}{%
\begin{minipage}{\dimexpr\linewidth-2\fboxsep-2\fboxrule\relax}
\textbf{Query:} If the spare bedroom AC is still on, switch it to dehumidify mode and lower the set temperature to around 21\,°C.
\end{minipage}}
\par\noindent
\begin{tcolorbox}[
    colframe=red,
    boxrule=0.85pt,
    colback=yellow!6!white,
    sharp corners,
    arc=0pt,
    boxsep=3pt,
    left=4pt,
    right=4pt,
    width=\linewidth
]
\textbf{Condition:}
\begin{lstlisting}[
language=Python,
columns=fullflexible,
keepspaces=true,
breaklines=true,
breakatwhitespace=false,
showstringspaces=false,
xleftmargin=0pt
]
(device("e7J6mKqC").ac_mode == "dry") and 
(20 <= device("e7J6mKqC").target_temperature <= 21)
\end{lstlisting}
\end{tcolorbox}

\textbf{Devices:} e7J6mKqC

\end{enumerate}

\end{tcolorbox}

\subsection{Dialogue Generation}

\subsubsection{Simulation Framework}

We construct a closed-loop interaction system consisting of, this enables grounded, executable, and stateful dialogue trajectories. Shown in Table~\ref{tab:environment-components}. 

\begin{table}[htbp]
\centering
\caption{Components in the interactive evaluation environment.}
\small
\begin{tabular}{ll}
\toprule
\textbf{Component} & \textbf{Role} \\
\midrule

Blueprint User (GPT-5) & Generates task-grounded user queries. \\
Assistant (GPT-5) & Produces responses and tool actions. \\
Sandbox & Executes actions and returns environment feedback. \\
Environment State & Dynamically updates after each interaction step. \\
\bottomrule
\end{tabular}
\label{tab:environment-components}
\end{table}

\subsubsection{Output Data Structure}

Each sample is represented in a unified structured format supporting both supervised and reinforcement learning. Shown in Table~\ref{tab:data_structure}.

\begin{table}[!htbp]
\centering
\caption{Unified data sample structure.}
\label{tab:data_structure}
\resizebox{\linewidth}{!}{%
\begin{tabular}{p{3.5cm} p{4.2cm} p{6.2cm}}
\toprule
\textbf{Component} & \textbf{Fields} & \textbf{Description} \\
\midrule

Meta Information 
& user profile, environment state, query type& Context conditioning signals. \\

Dialogue Messages 
& user input, assistant response, environment observation 
& Multi-turn interaction trajectory. \\

Task Status 
& completion flag 
& Execution-level supervision signal. \\

\bottomrule
\end{tabular}%
}
\end{table}

\subsection{Core Prompt Design}
\begin{tcolorbox}[
    breakable,
    width=\linewidth,
    colback=yellow!5!white,
    colframe=blue!60!black,
    title={\textbf{Blueprint Generation Prompt}},
    sharp corners,
    boxrule=0.6pt
]
\small

\textbf{Task Definition.}  
You are an expert system for generating smart home interaction scenarios grounded in real household environments and user profiles.

\textbf{Objective.}  
Synthesize realistic life scenes and corresponding atomic executable tasks for smart home systems, ensuring physical feasibility, behavioral realism, and strict device-level grounding.

\textbf{Generation Procedure.}

\textbf{1. Life Scene Construction}
\begin{itemize}
    \item Generate a fixed number of daily-life scenes.
    \item Each scene must describe continuous human behavior in a single physical location.
    \item Scenes must reflect user profile characteristics and environment states.
    \item Device states must be grounded in observed environment information.
    \item Do not generate command-style or task-list-style descriptions.
\end{itemize}

\textbf{2. Atomic Task Decomposition}
\begin{itemize}
    \item Decompose each scene into atomic tasks.
    \item Each task must originate from exactly one scene.
    \item Each task must represent a single user intent only.
    \item No new devices or attributes may be introduced.
\end{itemize}

\textbf{Hard Constraints.}

\textbf{1. Device Constraint}
\begin{itemize}
    \item Only devices explicitly defined in the environment are allowed.
    \item Device identifiers must match exactly.
\end{itemize}

\textbf{2. Attribute Constraint}
\begin{itemize}
    \item Only attributes listed in device service definitions may be used.
    \item No inferred or hallucinated attributes are allowed.
\end{itemize}

\textbf{3. Atomicity Constraint}
\begin{itemize}
    \item Each task must contain exactly one user intent.
    \item Multi-intent or composite tasks are not allowed.
\end{itemize}

\textbf{4. Physical Consistency Constraint}
\begin{itemize}
    \item All actions must be physically executable in the given environment.
    \item Device state transitions must be realistic and valid.
\end{itemize}

\textbf{5. Device State Dependency Constraint}
\begin{itemize}
    \item Functional attributes (e.g., temperature, brightness) are only valid when device is active.
    \item If device is inactive, activation must be included before modifying dependent attributes.
\end{itemize}

\textbf{Output Format.} Strict JSON only, no explanation or additional text.

\end{tcolorbox}
\vspace{1cm}
\begin{tcolorbox}[
    breakable,
    width=\linewidth,
    colback=yellow!5!white,
    colframe=blue!60!black,
    title={\textbf{User Profile Generation Prompt}},
    sharp corners,
    boxrule=0.6pt
]
\small

\textbf{Task Definition.} You are an expert system for generating smart home user profiles.

\textbf{Objective.} Synthesize realistic, diverse, and internally consistent user profiles for smart home automation systems.

\textbf{Design Principles.}

\textbf{1. Smart Home Grounding}
All attributes must map to controllable system behaviors:
\begin{itemize}
    \item Physiology $\rightarrow$ environment adaptation
    \item Health $\rightarrow$ HVAC / air quality control
    \item Habits $\rightarrow$ automation scheduling
    \item Preferences $\rightarrow$ scene control policies
\end{itemize}

\textbf{2. Completeness Constraints}
\begin{itemize}
    \item All modules must be present: basic, physiology, health, habit, preference
    \item No missing or null fields
    \item Full internal consistency required
\end{itemize}

\textbf{3. Realism Constraints}
\begin{itemize}
    \item Cover full demographic spectrum
    \item Include realistic health conditions
    \item Ensure physically plausible values
\end{itemize}

\textbf{Output Format.} Strict JSON only, no explanation.

\end{tcolorbox}


\noindent
\begin{tcolorbox}[
    breakable,
    width=\linewidth,
    colback=yellow!5!white,
    colframe=blue!60!black,
    title={\textbf{User Query Generation Prompt}},
    sharp corners,
    boxrule=0.6pt
]
\small
\textbf{Task Definition.} You are an expert system for generating natural user requests in a smart home environment.

\textbf{Objective.} Convert structured task states into realistic, minimal, and executable user queries that can be directly used for device control via tool invocation.

\textbf{Design Principles.}

\textbf{1. Task-Centric Grounding}
All outputs must strictly align with the current task list:
\begin{itemize}
    \item Focus only on \textbf{unfinished tasks}
    \item Ignore historical dialogue unless task-relevant
    \item Ensure every query directly contributes to task completion
\end{itemize}

\textbf{2. Device-State Awareness}
Generation must consider real-time home state:
\begin{itemize}
    \item Room-level and device-level status constraints
    \item Avoid redundant or already-satisfied operations
    \item Adjust actions based on current device conditions
\end{itemize}

\textbf{3. Natural User Expression Constraint}
Ensure outputs reflect realistic user behavior:
\begin{itemize}
    \item Avoid enumerating multiple parameters or rooms in a rigid format
    \item Avoid overly structured or machine-like command patterns
    \item Use short, conversational, and implicit instructions
\end{itemize}

\textbf{4. Minimalism Principle}
\begin{itemize}
    \item One unified query per generation step
    \item No task IDs, no explanations, no reasoning traces
    \item Merge actions only when naturally expressed
\end{itemize}

\textbf{5. Tool-Execution Requirement}
\begin{itemize}
    \item Output must be passed through \texttt{post\_process} tool
    \item Must include:
        \begin{itemize}
            \item \texttt{query}: a single executable instruction string
            \item \texttt{reason}: structured justification of generation logic
        \end{itemize}
    \item \texttt{query} must never be empty or null
\end{itemize}

\textbf{Output Format.} Strict tool-call format only. No free-text response.

\end{tcolorbox}
\begin{tcolorbox}[
    width=\linewidth,
    colback=yellow!5!white,
    colframe=blue!60!black,
    title={\textbf{Smart Home Agent Prompt}},
    sharp corners,
    boxrule=0.6pt
]
\small

\textbf{Task Definition.} You are an intelligent smart home agent with strong reasoning and device control capabilities. You operate inside a user’s home environment.

\textbf{Objective.} Accurately understand user intent (query/control), interact with smart home devices via tools, and return correct, executable results.

\textbf{System Context.}
\begin{itemize}
    \item \textbf{Home Devices:} Structured multi-room device list with metadata (id, did, category, brand, tags).
    \item \textbf{User Memory:} Historical preferences, behaviors, and contextual clips (if available).
    \item \textbf{User Scene Context:} Environmental and situational signals for personalization.
\end{itemize}

\textbf{Core Workflow (Strict).}
\begin{enumerate}
    \item \textbf{Intent Recognition:} Identify query type (status / control / ambiguous / chit-chat).
    \item \textbf{Information Retrieval:} Query device manuals and states before executing actions.
    \item \textbf{Disambiguation:} Apply clarification rules when intent or target is not deterministic.
    \item \textbf{Tool Execution:} Execute device operations only after validating capability and parameters.
    \item \textbf{Result Verification:} Check tool outputs; iterate if unmet.
    \item \textbf{User Response:} Return concise ($\leq 50$ Chinese characters), spoken-style answer only.
\end{enumerate}

\textbf{Clarification Principle.}
\begin{itemize}
    \item No clarification if device + function are uniquely identifiable.
    \item Ask user only when multiple valid interpretations exist.
    \item Avoid interpreting “all devices” unless explicitly stated (e.g., “all lights”).
\end{itemize}

\textbf{Execution Constraint.}
\begin{itemize}
    \item Always fetch device capability before control.
    \item Always ensure parameter validity before execution.
    \item Never output tool reasoning or internal steps to user.
\end{itemize}

\end{tcolorbox}

\section{MCTS-Flow for Dialogue Generation}
\vspace{-1cm}
In this section, we present an example of using MCTS-Flow to generate and select dialogue data.

\begin{figure}[t]
    \centering
    \includegraphics[width=0.9\linewidth]{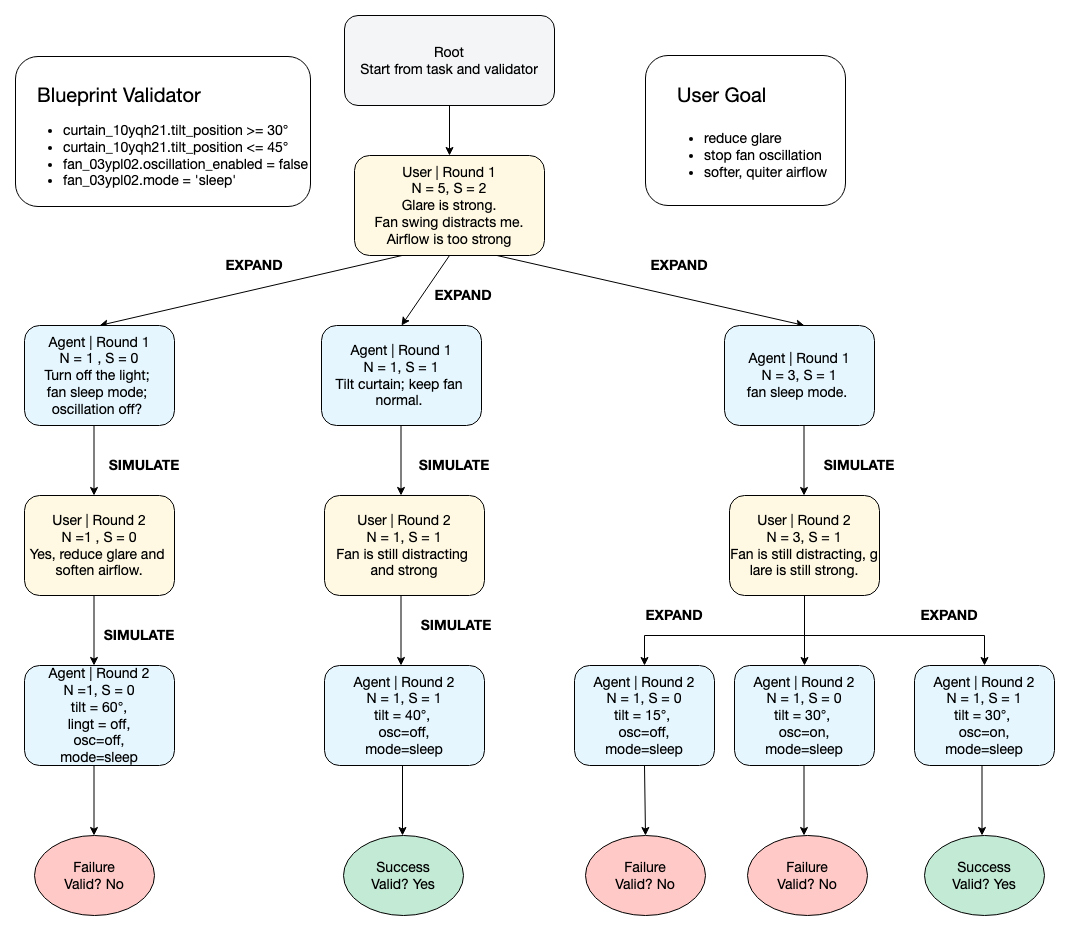}
    \caption{A demonstration of MCTS-Flow for dialogue generation.}
    \label{fig:mcts_demo}
\end{figure}

As illustrated in Figure~\ref{fig:mcts_demo}, the blueprint specifies the user goal, task objectives, and success metrics, collectively referred to as the User Goal and Blueprint Validator. A dialogue session is considered successful only if all conditions of the Blueprint Validator are satisfied upon termination.

In this example, MCTS-Flow performs 5 dialogue rollouts from a fixed blueprint. The initial user query is generated by the root node, followed by the agent's responses. For each user query, the agent generates at most 3 distinct candidate responses.

Some dialogues fail to meet the conditions of the Blueprint Validator upon termination and are thus deemed unsuccessful, as indicated in red. Two of them satisfy the conditions, as indicated in green. These successful instances can be collected to train the smart home agent. Notably, the successful ones share part of their dialogue trajectories with some of the failed ones.

As for this example, MCTS-Flow will collect 2 successful dialogue trajectories, as shown in the following 2 figures. Note that these 2 dialogues meet the user's requirements in different ways, enhancing data diversity.

\begin{center}
    \begin{tcolorbox}[colback=yellow!6!white, colframe=blue!50!green, title=\textbf{Success Dialogue 1}]
    \small
    \textbf{User Query:} "The glare is strong, the fan swing distracts me. And the airflow is too strong."
    
    \medskip
    \textbf{Smart Home Agent:}
    \begin{itemize}
        \item \texttt{[Query]} $\rightarrow$ Query the device status of the room.
        \item \texttt{[Device Control]} $\rightarrow$ Tilt the curtain to 40 degree.
    \end{itemize}
    \medskip
    \textbf{User Query:} "The fan is still distracting me. And the airflow is too strong."
    
    \medskip
    \textbf{Smart Home Agent:} 
    \begin{itemize}
        \item \texttt{[Query]} $\rightarrow$ Query the device specification of the room.
        \item \texttt{[Device Control]} $\rightarrow$ Set the oscillation of fan to 'off'. Set the mode of the fan to 'sleep'
    \end{itemize}
    \end{tcolorbox}
    \label{fig:success_sampe_1}
\end{center}

\begin{center}
    \begin{tcolorbox}[colback=yellow!6!white, colframe=blue!50!green, title=\textbf{Success Dialogue 2}]
    \small
    \textbf{User Query:} "The glare is strong. The fan swing distracts me. And the airflow is too strong."

    \medskip
    \textbf{Smart Home Agent:}
    \begin{itemize}
        \item \texttt{[Query]} $\rightarrow$ Query the device status of the room.
        \item \texttt{[Device Control]} $\rightarrow$ Set the mode of the fan to 'sleep'.
    \end{itemize}
    \medskip
    \textbf{User Query:} "The fan is still distracting me. And the glare is still strong."

    \medskip
    \textbf{Smart Home Agent:} 
    \begin{itemize}
        \item \texttt{[Query]} $\rightarrow$ Query the device specification of the room.
        \item \texttt{[Device Control]} $\rightarrow$ Set the oscillation of fan to 'off'. Tilt the curtain to 40 degree.
    \end{itemize}
    \end{tcolorbox}
    \label{fig:success_sampe_2}
\end{center}

\section{Details of the Two-Stage Training} 

\subsection{Training Data Volume and Composition}

The supervised fine-tuning (SFT) dataset consists of three distinct subsets, all derived from a shared foundational pool of 20,000 user profiles. Specifically, we first generated these 20,000 profiles, each encapsulating attributes such as age, occupation, and behavioral habits. Subsequently, HomeMaker was employed to procedurally generate corresponding household configurations for each user. Building upon this foundation, dialogue trajectories were synthesized corresponding to the three data generation schemes evaluated in our ablation study:

\textbf{Pure-Text Role-Play (w/o Blueprint):} Based on the 20,000 user profiles with basic household information, we adopted a conventional pure-text role-playing scheme without strict physical blueprint constraints. This initial process produced 20,000 dialogues. After applying an LLM-based filter to ensure quality, 16,700 dialogues were retained, from which 15,000 samples were randomly selected to form the first subset.

\textbf{Blueprint-Grounded Linear Generation (w/o MCTS-Flow):} To incorporate physical grounding, HomeMaker was utilized to generate explicit household blueprints for the same 20,000 users. Following the linear generation scheme (without search-based exploration), we obtained 17,500 physically grounded dialogue trajectories, from which 15,000 samples were randomly selected for this subset.

\textbf{Blueprint-Grounded MCTS-Flow (Complete Pipeline):} Applying our complete data synthesis pipeline, we coupled the blueprint configurations with MCTS-Flow search. Guided by verification signals, this dynamic exploration phase yielded a highly diverse set of 30,000 verifiable dialogues. From these, 15,000 samples were randomly selected to form the final, highest-quality SFT subset.

For the subsequent reinforcement learning (RL) stage, the training environments are initialized from a distinct set of 10,000 base configurations, each encompassing a user profile, household information, and a physical blueprint. We constructed two corresponding setups to support our online policy optimization:

\textbf{HomeFlow-RL (Complete Step-wise RLVE):} The 10,000 base configurations were utilized directly to initialize the HomeEnv simulator, enabling the agent to engage in complete, multi-turn episodic interactions guided by step-wise dense rewards.

\textbf{HomeFlow-RL (ST):} To construct the single-turn ablation baseline, we first generated complete multi-turn trajectories from the same 10,000 base configurations. We then stripped away all intermediate tool-use and device control turns, retaining only the final single-turn interaction. This limits the agent to perceiving the full context but only receiving sparse feedback based on the ultimate outcome.

\subsection{Training Setups}

\textbf{Reward Shaping:} We design three rule-based reward components to constrain model behavior. 

\textbf{CodeCheck} penalizes the use of restricted Python operations, including introspection and dynamic execution-related keywords such as \textit{getattr}, \textit{setattr}, \textit{eval}, \textit{exec}, and \textit{compile}, among others. 

\textbf{CodeLength} enforces a maximum code length of 1000 tokens, assigning a reward of $-1$ if exceeded and $0$ otherwise. 

\textbf{TurnCheck} limits the interaction horizon to 3 turns, similarly assigning a penalty of $-1$ when violated and $0$ otherwise.

The detailed hyperparameters for both HomeFlow-RL setups are summarized in Table~\ref{tab:st-rl-settings}.

\begin{table}[htbp]
\centering
\captionsetup{skip=6pt}
\caption{RLVE training hyperparameters. Bold values mark the key divergence: HomeFlow-RL (ST) employs entropy regularization ($\beta_{\text{entropy}}=10^{-4}$) to combat collapse, while HomeFlow-RL (Full) removes entropy loss and instead enforces a KL constraint ($\beta_{\text{KL}}=3\times 10^{-3}$) against the SFT reference, relying on the dynamic User simulator for natural exploration.}
\label{tab:st-rl-settings}

\begin{tabular}{lcc}
\toprule
\textbf{Parameters} & \textbf{HomeFlow-RL (ST)} & \textbf{HomeFlow-RL (Full)} \\
\midrule
\textbf{GPU Type} & \multicolumn{2}{c}{Nvidia H20 96GB} \\
\textbf{Memory} & \multicolumn{2}{c}{1TB} \\
\textbf{Base Model} & \multicolumn{2}{c}{Qwen3-8B-SFT} \\
\textbf{TP} & \multicolumn{2}{c}{4} \\
\textbf{PP} & \multicolumn{2}{c}{1} \\
\textbf{CP} & \multicolumn{2}{c}{1} \\
\textbf{SP} & \multicolumn{2}{c}{on} \\
\textbf{Rollout group size} & \multicolumn{2}{c}{8} \\
\textbf{Rollout temperature} & \multicolumn{2}{c}{1.0} \\
\textbf{LR} & \multicolumn{2}{c}{1e-6} \\
\textbf{LR warmup fraction} & \multicolumn{2}{c}{0.01} \\
\textbf{Minimize LR} & \multicolumn{2}{c}{5e-7} \\
\textbf{TaskFinish weight} & \multicolumn{2}{c}{1.0} \\
\textbf{CodeCheck weight} & \multicolumn{2}{c}{0.2} \\
\textbf{CodeLength weight} & \multicolumn{2}{c}{0.2} \\
\textbf{TurnCheck weight} & \multicolumn{2}{c}{0.2} \\

\midrule
\textbf{Num of GPUs} & \textbf{32} & \textbf{96} \\
\textbf{Training time cost} & \textbf{38h+} & \textbf{95h+} \\
\textbf{KL Loss coefficient} & \textbf{0.0} & \textbf{3e-3} \\
\textbf{Entropy Loss coefficient} & \textbf{1e-4} & \textbf{0.0} \\
\textbf{User LLM} & -- & \textbf{DeepSeek-V3.2} \\

\bottomrule
\end{tabular}
\end{table}

\section{SmartHome-Bench}
\label{app:dataset}
We constructed SmartHome-Bench, a comprehensive benchmark featuring 1,678 test samples systematically categorized into 5 major groups and 18 subcategories (see Table \ref{tab:user_template_rules}). To rigorously test the agent's scalability and spatial reasoning, the environmental difficulty is highly diversified, encompassing a broad distribution of household layouts ranging from simple setups with fewer than 10 devices to highly complex homes with over 100 devices.

\begingroup
\small
\setlength{\tabcolsep}{3pt}
\begin{longtable}{>{\raggedright\arraybackslash}p{0.20\linewidth}
                  >{\raggedright\arraybackslash}p{0.23\linewidth}
                  >{\raggedright\arraybackslash}p{0.29\linewidth}
                  >{\raggedright\arraybackslash}p{0.22\linewidth}}
\caption{Template families and generation rules.}
\label{tab:user_template_rules}\\
\toprule
Family & Templates & Grounding rule & Surface instruction rule \\
\midrule
\endfirsthead
\caption[]{Template families and generation rules (continued).}\\
\toprule
Family & Templates & Grounding rule & Surface instruction rule \\
\midrule
\endhead
\multicolumn{4}{r}{Continued on next page} \\
\midrule
\endfoot
\bottomrule
\endlastfoot

\texttt{TC1: Atomic}\newline
\texttt{Control} &
\texttt{Clear Commands};\newline
\texttt{Colloquial Requests};\newline
\texttt{Noisy Utterances} &
Select one controllable device and one supported service or attribute change. The target must be unique and directly executable. &
\texttt{clear} is explicit; \texttt{colloquial} uses everyday phrasing; \texttt{noisy} may include filler words, repetition, ASR-like artifacts, or immediate self-correction. \\
\midrule
\texttt{TC2: Compositional}\newline
\texttt{Control} &
\texttt{Explicit Multi-device};\newline
\texttt{Batch Operations};\newline
\texttt{State-dependent};\newline
\texttt{Room-dependent};\newline
\texttt{Top-N Selection} &
Select multiple operations, a device set, a state-filtered subset, a room-conditioned subset, or a ranked top-$k$ target. All selected devices must support the requested operation. &
Express the dependency in one household request, e.g., ``all lights'', ``lights above 80\%'', ``rooms where the AC is on'', or ``the two fastest fans''. \\
\midrule
\texttt{TC3: Ambiguous}\newline
\texttt{Intent} &
\texttt{State-level Ambiguity};\newline
\texttt{Scene-level Ambiguity};\newline
\texttt{Slot-missing} &
Choose a plausible target but make the initial utterance under-specified, subjective, or scene-like. Missing slots may include room, device, mode, value, duration, or time. &
The initial query should require clarification or context-based inference. The hidden target is preserved through \texttt{intended\_query} or the reference clarification trajectory. \\
\midrule
\texttt{TC4: Context-Aware}\newline
\texttt{Multi-turn}\newline
\texttt{Interaction} &
\texttt{Context Continuation};\newline
\texttt{Topic Shifts};\newline
\texttt{Coreference Resolution};\newline
\texttt{User Corrections} &
The current target depends on previous turns. The generator may inherit a device, override a previous operation, resolve a pronoun, or switch topics while preserving context. &
Use natural dialogue forms such as ``also do that'', ``no, make it 26'', ``turn that one off'', or a topic-shifted follow-up. \\
\midrule
\texttt{TC5: Personalized}\newline
\texttt{Memory} &
\texttt{Short-term Memory};\newline
 &
Use recent interaction facts or stored user preferences to fill omitted action details. The task should not be solvable in the intended way without the provided memory. &
Allow compact preference-triggering instructions such as ``sleep time'', ``as usual'', or ``set it how I like it''. \\
\end{longtable}
\endgroup

\FloatBarrier

\section{Case Study of the Impact of RL Training on Model Ability}
\label{app:case_study}

In this section, we present several case studies on RL training. Case 1 illustrate the improvement in the model’s comprehension ability after RL training. Case 2 illustrate the improvement in the model’s prioritization skill ability and reasoning ability after RL training. 

\begin{center}
    \begin{tcolorbox}[colback=yellow!6!white, colframe=blue!50!green, title=\textbf{Case 1(Before RL): Comprehension Ability}]
    \small
    \textbf{User Query:} "Adjust the living room air conditioner to 20 degree and enable vertical swing."

    \medskip
    \textbf{Smart Home Agent:} "The living room air conditioner has been set to 20 degree with vertical swing enabled."

    \medskip
    \textbf{User Query:} "Do the same for the master bedroom and guest bedroom, and open all the curtains as well."

    \medskip
    \textbf{Smart Home Agent (Failed):}
    \begin{itemize}
        \item \texttt{[Query]} $\rightarrow$ Query the device specification of the curtains.
        \item \texttt{[Device Control]} $\rightarrow$ Open the curtains.
        \item \texttt{[Response]} $\rightarrow$ The master bedroom curtain and guest bedroom curtain have both been opened.
        \item \textit{Outcome:} The assistant doesn't set the air conditioners for both the master bedroom and the guest bedroom.
    \end{itemize}
    \end{tcolorbox}
    \label{fig:case_1_before_RL}
\end{center}

\begin{center}
    \begin{tcolorbox}[colback=yellow!6!white, colframe=blue!50!green, title=\textbf{Case 1(After RL): Comprehension Ability}]
    \small
    \textbf{User Query:} "Adjust the living room air conditioner to 20 degree and enable vertical swing."
    \medskip
    \textbf{Smart Home Agent:} "The living room air conditioner has been set to 20 degree with vertical swing enabled."
    
    \medskip
    \textbf{User Query:} "Do the same for the master bedroom and guest bedroom, and open all the curtains as well."

    \medskip
    \textbf{Smart Home Agent (Succeed):}
    \begin{itemize}
        \item \texttt{[Query]} $\rightarrow$ \textbf{Query the device specification of the air conditioners.}
        \item \texttt{[Device Control]} $\rightarrow$ \textbf{Set the air conditioners.}
        \item \texttt{[Query]} $\rightarrow$ Query the device specification of the curtains.
        \item \texttt{[Device Control]} $\rightarrow$ Open the curtains.
        \item \texttt{[Response]} $\rightarrow$ \textbf{The master bedroom and guest bedroom air conditioners have both been turned on, set to 20 degree, and vertical swing has been enabled.} The master bedroom curtain and guest bedroom curtain have both been opened.
        \item \textit{Outcome:} The assistant completed both the air conditioner and curtain actions, correctly understanding "do the same".
    \end{itemize}
    \end{tcolorbox}
    \label{fig:case_1_after_RL}
\end{center}

\begin{center}
    \begin{tcolorbox}[colback=yellow!6!white, colframe=blue!50!green, title=\textbf{Case 2(Before RL): Prioritization Skill Ability and Reasoning Ability}]
    \small
    \textbf{User Query:} "It feels a little cool. Change the air conditioner to natural breeze."
    
    \medskip
    \textbf{Smart Home Agent:} "There are air conditioners running in multiple rooms. Which room would you like to adjust?"

    \medskip
    \textbf{User Query:} "The master bedroom."

    \medskip
    \textbf{Smart Home Agent (Failed):}
    \begin{itemize}
        \item \texttt{[Device Control]} $\rightarrow$ Try to set the mode of the air conditioner to 'natural'.
        \item \textit{Outcome:} The air conditioner has no "natural" mode thus this command is failed. 
    \end{itemize}
    \end{tcolorbox}
    \label{fig:case_2_before_RL}
\end{center}

\begin{center}
    \begin{tcolorbox}[colback=yellow!6!white, colframe=blue!50!green, title=\textbf{Case 2(After RL): Prioritization Skill Ability and Reasoning Ability}]
    \small
    \textbf{User Query:} "It feels a little cool. Change the air conditioner to natural breeze."
    
    \medskip
    \textbf{Smart Home Agent:} "There are air conditioners running in multiple rooms. Which room would you like to adjust?"

    \medskip
    \textbf{User Query:} "The master bedroom."

    \medskip
    \textbf{Smart Home Agent (Succeed):}
    \begin{itemize}
        \item \texttt{[Query]} $\rightarrow$ \textbf{Query the device specification of the air conditioner in master bedroom.}
        \item \texttt{[Reason]} $\rightarrow$ \textbf{Find that there is no 'natural' mode, and guess that the user want the 'fan\-only' mode.}
        \item \texttt{[Device Control]} $\rightarrow$ \textbf{Set the mode of the air conditioner to 'fan\-only'.}
        \item \textit{Outcome:} Correctly reads the device status and reasonably maps 'natural breeze' to 'fan-only' mode. 
    \end{itemize}
    \end{tcolorbox}
    \label{fig:case_2_before_RL}
\end{center}

%

\end{document}